\documentclass[11pt]{article}

\usepackage[]{acl}

\usepackage{times}
\usepackage{latexsym}
\usepackage{amsmath, amssymb, mathtools}
\usepackage{multirow}
\usepackage{graphicx}
\usepackage{algorithm}
\usepackage{algpseudocode}
\usepackage{subcaption}
\usepackage{enumitem}
\usepackage{tcolorbox}
\usepackage{booktabs}

\usepackage{comment}
\usepackage[T1]{fontenc}

\usepackage[utf8]{inputenc}

\usepackage{microtype}

\usepackage{inconsolata}

\usepackage{graphicx}

\title{Detecting Non-Membership in LLM Training Data via Rank Correlations}

\author{Pranav Shetty, Mirazul Haque, Zhiqiang Ma, Xiaomo Liu \\
        JPMorgan AI Research \\ \texttt{first.last@jpmchase.com}}

\newcommand{\method}{\textsc{PRISM}}
\newcommand{\data}{\(\mathcal{D}\)}
\newcommand{\target}{\(M_{T}\)}
\newcommand{\reference}{\(M_{R}\)}
\newcommand{\distilled}{\(M_{D}\)}

\newcommand{\minkpp}{Min-K\%++ }

\begin{document}

\maketitle

\begin{abstract}

As large language models (LLMs) are trained on increasingly vast and opaque text corpora, determining which data contributed to training has become essential for copyright enforcement, compliance auditing, and user trust.
While prior work focuses on detecting whether a dataset \emph{was} used in training (membership inference), the complementary problem—verifying that a dataset was \emph{not} used- has received little attention.
We address this gap by introducing \method{}, a test that detects dataset-level \emph{non-membership} using only grey-box access to model logits.
Our key insight is that two models that have not seen a dataset exhibit higher rank correlation in their normalized token log probabilities than when one model has been trained on that data.
Using this observation, we construct a correlation-based test that detects non-membership.
Empirically, \method{} reliably rules out membership in training data across all datasets tested while avoiding false positives, thus offering a framework for verifying that specific datasets were excluded from LLM training.



\end{abstract}

\section{Introduction}

Large Language Models (LLMs) owe their broad general-purpose capabilities to training on massive, diverse corpora of high-quality text \cite{wettig2024qurating, zmigrod-etal-2024-value}. Much of this material is scraped from the open internet, raising complex legal and compliance issues for every stakeholder in the ecosystem—from content creators and publishers to model developers and end-users. Many publishers have filed legal challenges that their content has been used without authorization and compensation to train LLMs and face the burden of proving that their data has been used for training \cite{brittain_authors_sue_anthropic_2024, reuters_nytimes_openai_2023}. On the flip side, LLM trainers face the opposite burden of proving that they have \emph{not} used a given dataset for training. Similarly, enterprises onboarding LLMs for internal use must perform compliance audits of these models to understand the legal risk in using these models in case the models have been trained using proprietary datasets \cite{t-y-s-s-etal-2025-cocolex}. Similar verification is essential for enterprises that adopt third-party LLMs, which must audit models for potential exposure to proprietary or restricted data to assess legal and compliance risk. Non-membership detection is equally critical when users or organizations wish to ensure that sensitive or legally protected datasets—shared under agreements restricting their use for training—have genuinely been excluded. As modern LLMs are increasingly trained on corpora approaching “the sum of all human text,” the ability to reliably establish what a model has not seen becomes as important as identifying what it has. 

Membership inference attacks (MIA) have been studied in the literature to assess whether a certain dataset has been used to train a model (i.e., is a member of the training data) \cite{shi2023detecting, zhang-etal-2024-pretraining}. These MIA scores typically rely on analyzing the token probabilities and comparing the token probabilities of member data (i.e., used for training) against non-member data (not used during training) from the same distribution. Member data typically has a higher probability than similar data not seen during training. However, these methods were shown to overfit to distributional differences between member and non-member data and performed no better than random once these differences were corrected \cite{duan2024membership}. To address this issue, LLM Dataset Inference (DI) was proposed \cite{mainidi2024}. LLM-DI does not perform inference on each document in a dataset but instead performs inference on the entire dataset. As this reduces variance due to outliers, we adopt this setting in our work. 

However, LLM-DI and previously proposed MIA techniques have two major limitations. Firstly, they require access to known non-member data from an identical distribution to detect membership. In practice, obtaining such a distributionally matched reference dataset is rarely feasible, especially for proprietary or domain-specific corpora. Second, existing methods are inherently one-sided: they are designed only to provide evidence for membership in the training data.
To date, no method to the best of our knowledge explicitly addresses the complementary problem of non-membership detection, i.e., ruling out that a dataset was used during training. Methods like LLM-DI perform a one-tailed hypothesis test and can either detect that a dataset was used for training or fail to detect this. Mere failure to detect membership does not automatically imply evidence of non-membership. This asymmetry leaves open a crucial gap: how to confidently demonstrate that a model has not been trained on a given dataset.




To address these limitations, we propose Normalized token log \textbf{P}robability \textbf{R}ank correlation \textbf{I}nference using \textbf{S}pearman for non-\textbf{M}embership detection (\method{}). We use the \minkpp score as our signal for detecting membership. \minkpp computes the average of the lowest $K$ token log probabilities in any document, normalized using the mean and variance over the model's vocabulary. We begin with a reference model known not to have been trained on the suspect dataset and a target model whose training history we wish to test.
Our central insight is that the Spearman rank correlation between the \minkpp{} scores produced by these two models serves as a reliable indicator of dataset membership.
If the target model has not been trained on the dataset, both models—being sufficiently capable language models—will rank documents according to their underlying linguistic difficulty, yielding a high correlation. Conversely, if the target model has been trained on the dataset, memorization effects distort this ranking, leading to a lower correlation.  

Our core contributions are as follows:

\begin{enumerate}
    \item We propose the task of dataset-level non-member detection and demonstrate that \method{} reliably detects non-membership across all datasets and models using only the logits of the model.
    \item \method{} does not require access to any additional reference datasets that were known to be held out from training a model, which is required by existing methods and is challenging to obtain in practice \cite{zhaounlocking}.
    \item \method{} can be performed with as few as 100 documents, making it practical to use.
\end{enumerate}

\begin{figure*}[t]
    \centering
    \includegraphics[width=0.9\textwidth]{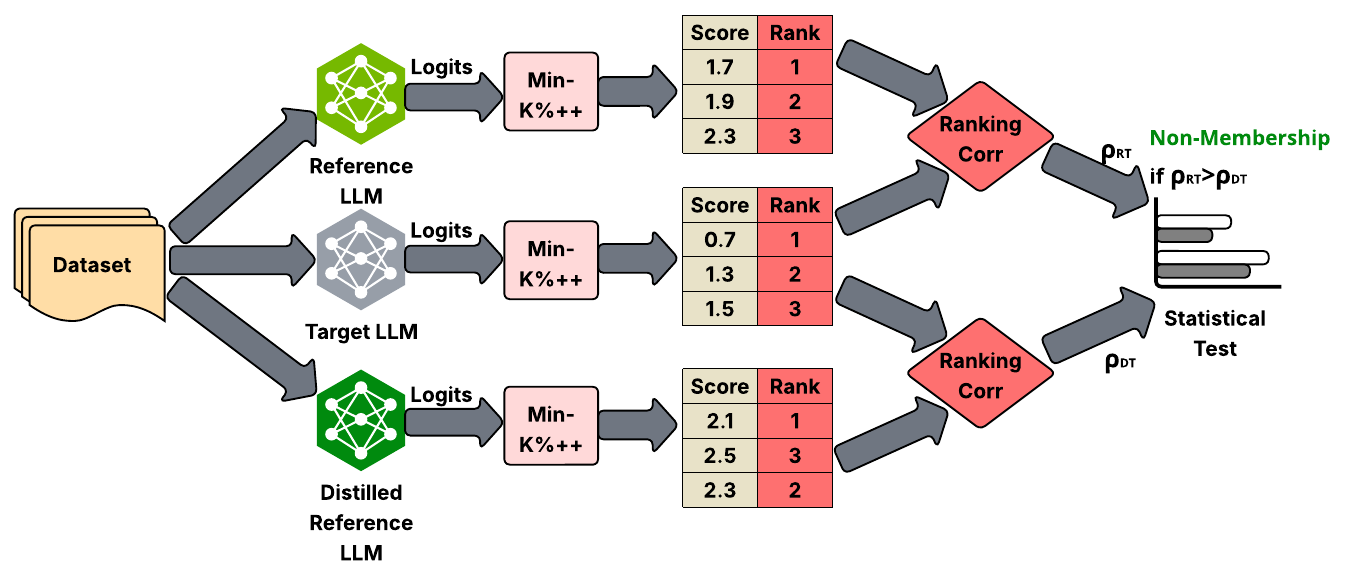}
    \caption{Overview of \method{}. Logits of Reference LLM, Target LLM, and Distilled Reference LLMs are used to calculate \minkpp{} scores, and the corresponding ranking of the scores. Final statistical testing for non-membership detection is dependent on the rank correlation between the reference and target model, and the rank correlation between the distilled reference and target model.}
    \label{fig:method}
\end{figure*}

\section{Related Work}

\noindent\textbf{Membership Inference Attack}. Early research on membership inference attack (MIA) focused on simpler neural architectures and convolutional neural networks, and involved training multiple shadow models on various dataset partitions to extract features indicative of membership status \cite{shokri2017membership}. However, this shadow model approach does not scale to LLMs due to their large parameter count and computational demands. To address this limitation, more recent work has shifted towards using the log probability scores produced by LLMs to distinguish between member and non-member data points \cite{zhang-etal-2024-pretraining, shi2023detecting}. 

To facilitate the evaluation of these log-probability-based MIA techniques, researchers introduced benchmark datasets such as WikiMIA \cite{shi2023detecting} and PatentMIA \cite{zhang-etal-2024-pretraining}. These datasets were curated by selecting documents published before the LLM’s training cutoff as members and those published after as non-members. However, subsequent analysis revealed that the effectiveness of these attacks was largely attributable to temporal artifacts inherent in the dataset construction process. Temporal shifts lead to shifts in the specific language used in the member and non-member datasets, which is what these methods detected.
When member and non-member data were drawn from the same distribution, the performance of all evaluated MIA methods dropped to no better than random, indicating that prior methods did not infer membership but rather they exploited these temporal signals \cite{duan2024membership, mainidi2024, das2025blind}.

\noindent\textbf{Dataset Inference}.
Unlike traditional membership inference attacks, which aim to determine the membership status of individual data points, Dataset Inference (DI) shifts the focus to assessing the likelihood that an entire dataset was included in the training process \cite{mainidi2024}. Rather than producing granular, document-level predictions, DI seeks to estimate a confidence score for the entire dataset. This approach is motivated by the observation that LLM training pipelines often ingest data in bulk from specific sources, resulting in groups of related documents being consistently present in the training set. By aggregating signals across multiple documents, DI not only amplifies the overall detection capability but also mitigates the impact of noisy or anomalous samples \cite{shetty2026perturb}.

\section{Preliminaries}
\label{sec:prelim}
\subsection{Problem setup}

We consider the setting in which a verifier wishes to detect with high confidence whether a given suspect dataset \data{} was \emph{not} used in training a target LLM (non-membership detection), denoted by \target{}. \data{} is a collection of documents $\{x^{(i)}\}_{i=1}^n$, where each $x^{(i)} = (x^{(i)}_1, x^{(i)}_2, \ldots, x^{(i)}_{k_i})$ is a sequence of tokens drawn from a vocabulary $\mathcal{V}$.
We assume \emph{grey-box access} to \target{}, meaning that for each token $x_t$ in a sequence $x$, we can query the model to obtain its log probabilities over the known vocabulary $\mathcal{V}$. Formally, for a prefix $x_{<t}$, we may compute
\[
\ell_{M_T}(x_t \mid x_{<t}) = \log P(x_t \mid x_{<t}; M_T),
\]
where $P(\cdot \mid x_{<t}; M_T)$ denotes \(M_T\)'s predicted distribution over $\mathcal{V}$ at step $t$. Crucially, we do not assume access to model parameters or gradients. This level of access is consistent with what is available for some commercial LLM APIs. Using model logits is more informative than a black-box approach but less so than using deeper layers of the model \cite{chen-etal-2025-statistical}. Our choice of a grey-box approach is motivated by practical applicability rather than downstream performance, as access to target model weights is often not feasible.

\paragraph{Reference LLM.}
We assume access to a reference LLM (\reference{}), which is open-source and known not to have been trained on \data{}. Such a model can often be identified by leveraging release dates: if the suspect dataset’s public availability date is known, a model whose training data cutoff predates that release can be used as \reference{}. In practice, for many types of data, such as scientific literature, news articles, and online forums, establishing such a timestamp is feasible. 

\subsection{\minkpp}
Many scores have been proposed for membership detection, such as LOSS \cite{yeom2018privacy}, Min-k \cite{shi2023detecting}, and zlib \cite{carlini2021extracting} (Appendix \ref{app:mia_methods}).
\minkpp{}\citep{zhang2025mink} focuses specifically on the \(K\%\) of tokens that have the lowest normalized likelihood under a model. \minkpp incorporates normalization relative to the mean and variance of token log probabilities. 


Formally, given an autoregressive model \(M\) and a token sequence \( x = (x_1, x_2, \dots, x_n) \), define the token-level normalized log probability as 
\begin{equation*}
z(x_t; M) = \frac{\log P(x_t \mid x_{<t}; M) - \mu_{x_{<t}}}{\sigma_{x_{<t}}},
\end{equation*}
where
\begin{align*}
\mu_{x_{<t}} &= \mathbb{E}_{z \sim P(\cdot \mid x_{<t}; M)} [\log P(z \mid x_{<t}; M)],\\
\sigma_{x_{<t}} &= \sqrt{\mathrm{Var}_{z \sim P(\cdot \mid x_{<t}; M)} [\log P(z \mid x_{<t}; M)]}.
\end{align*}
Here, \(\mu_{x_{<t}}\) represents the expectation of the log probability distribution for the next token given the prefix \(x_{<t}\), and \(\sigma_{x_{<t}}\) denotes the corresponding standard deviation. \(\mathrm{Var}\) denotes the variance.

The \minkpp{} score for a sequence \(x\) is defined as the average of the normalized log probabilities \(z(x_t; M)\) over the \(K\%\) of tokens in the sequence with the lowest values (indicating highest surprisal):

\begin{multline*}
f_{\text{Min-K\%++}}(x; M) \\ = \frac{1}{|\text{Min-k}(x)|}\sum_{x_t \in \text{Min-K}(x)} z(x_t; M).
\end{multline*}
This normalization allows \minkpp to better distinguish sequences that are part of the training data from those that are not, by highlighting the relative surprisal of the most unlikely tokens while making it robust to absolute probability shifts across tokens.
Critically, prior work \citep{zhang2025mink} has shown that the \minkpp score theoretically corresponds to measuring the negative trace of the Hessian matrix of the log-likelihood \(log\; P(x_t \mid x_{<t}; M)\). Intuitively, training via maximum-likelihood directly reduces the curvature (Hessian trace) of the loss landscape at training examples, thereby causing their corresponding \minkpp scores to increase.

\subsection{Empirical motivation for \method{}}

To ground our approach, we begin with an empirical observation using the Pythia-410m model family.
We compute the \minkpp{} scores of documents under both the original Pythia-410m model and a version that has undergone continued pretraining on additional data, referred to as Pythia-410m-CPT.
Specifically, Pythia-410m-CPT was trained on several datasets drawn from the Pile validation split (listed in Table \ref{tab:dataset_correlations})—datasets that were not part of the original Pythia-410m training—and exposed to an additional six billion tokens of text.

We then measure the Spearman rank correlation of \minkpp{} scores between each model (Pythia-410m or Pythia-410m-CPT) and a set of independent reference models from the same Pythia family that have not seen these datasets.
Across all datasets and reference models, we observe a consistent pattern:
the rank correlation between the untrained Pythia-410m and each reference model is higher than that between the trained Pythia-410m-CPT and the same reference model.
Among several membership-inference metrics we tested, \minkpp{} produced the most consistent and separable differences (see Appendix \ref{app:correlation_trends}).

This pattern suggests an intuitive mechanism.
Two transformer models trained on large, overlapping web corpora but not exposed to the target dataset will tend to share similar priors about linguistic difficulty and token surprisal, leading them to rank documents in roughly the same order.
When one of these models is later trained on that dataset, memorization effects perturb these priors and alter the ranking, thereby lowering the rank correlation.

This observation motivates our hypothesis:
The relative rank correlation between a target model and a reference model can serve as a robust signal of dataset membership.
In particular, a higher correlation with a reference model that has not seen the data—compared to one that has—provides strong evidence of non-membership.
The next section formalizes this idea into a statistical test capable of detecting non-membership.


\begin{table}[htbp]
\centering
\setlength{\tabcolsep}{3pt}
\scriptsize
\caption{Rank correlation coefficients over different datasets between Pythia reference models of different sizes computed against Pythia-410m and Pythia-410m-CPT. Each of the datasets was held out from training the vanilla model as well as each reference model in the columns, but was used during continued pretraining of Pythia-410m-CPT with 6 billion additional tokens.}
\label{tab:dataset_correlations}
\begin{tabular}{l|cc|cc|cc}
\hline
Dataset & \multicolumn{2}{c|}{Pythia-1b} & \multicolumn{2}{c|}{Pythia-2.8b} & \multicolumn{2}{c}{Pythia-6.9b} \\
\cline{2-7}
& 410m & 410m-CPT & 410m & 410m-CPT & 410m & 410m-CPT \\
\hline
ArXiv  & 0.82 & 0.45 & 0.75 & 0.42 & 0.72 & 0.37 \\
HN & 0.71  & 0.40 & 0.74 & 0.40 & 0.68 & 0.41 \\
CC & 0.80 & 0.50 & 0.77 & 0.46 & 0.70 & 0.42 \\
Wikipedia & 0.84 & 0.25 & 0.80 & 0.25  & 0.77  & 0.21 \\
PubMed & 0.84 & 0.33 & 0.79 & 0.29 & 0.77 & 0.30 \\
\hline
\end{tabular}
\end{table}

\section{Methods}

\subsection{Comparing Rank Correlations}


\method{} operates by comparing how the suspect dataset is ranked, in terms of difficulty, across three models:
(1) the target model (\target{}),
(2) the reference model (\reference{}), and
(3) a derived model called the distilled reference (\distilled{}).
Figure~\ref{fig:method} illustrates this setup.

The distilled reference is created by fine-tuning the reference model on the suspect dataset while performing knowledge distillation from the target model.
This dual training process uses the target model as a teacher—aligning the logits of the distilled reference with those of the target—while simultaneously training it on the suspect data.
As a result, \distilled{} approximates how the target model would behave if it had been trained on the dataset, effectively serving as a proxy for “the target model exposed to \data{}.”

\method{} computes \minkpp{} scores for all documents in the dataset using each of the three models and evaluates their Spearman rank correlations.
If the target model has not been trained on the suspect data, its correlation with the reference model
will exceed its correlation with the distilled reference, since only the latter has been exposed to \data{}.
Conversely, if the target model has been trained on \data{}, it will align more closely with the distilled reference, resulting in higher correlation with it due to knowledge distillation. A crucial practical consideration is balancing the fine-tuning and distillation when training the distilled reference.
Excessive distillation can make the model too similar to the target, obscuring the difference in rank correlations.



We use rank correlations here as they are robust to changes in model scale between the target model and scoring model. They are also robust to outliers \cite{schober2018correlation, tabatabai2021introduction}.

\subsection{Training \distilled{}}

\begin{figure}[t]
    \centering
    \includegraphics[width=0.45\textwidth]{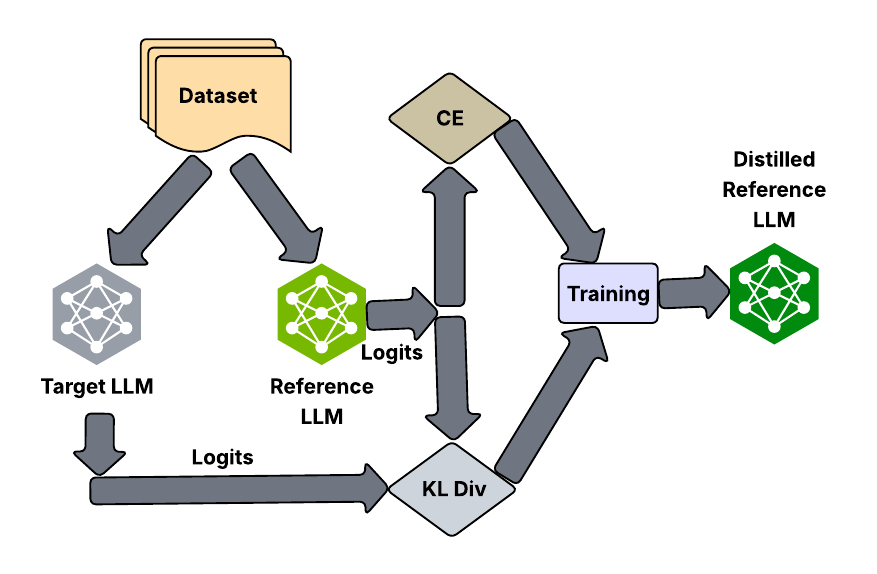}
    \caption{Overview of distilling the reference model. The distillation is dependent on two components: cross entropy of the reference model and KL divergence between the output probability of the target and reference model.}
    \label{fig:distillation}
\end{figure}

We obtain \distilled{} by simultaneously fine-tuning on \data{} and performing knowledge distillation using the logits of \target{} to minimize the Kullback-Leibler (KL) divergence \cite{kullback1951information} between the probability distributions of the target and reference models (Figure \ref{fig:distillation}). 

The loss used to obtain \distilled{} from \reference{} is a linear combination of a supervised cross-entropy loss on the true tokens and a distillation term encouraging \distilled{} to match the logits of \target{}. Formally, during training, we minimize the following loss over \data{}.


\[
\begin{split}
\mathcal{L}(\theta)
&= (1-\lambda) \,\mathbb{E}_{x\sim \mathcal{D}}\,\sum_{t=1}^{|x|}
    \underbrace{\mathrm{CE}\!\left(P_{M_{\mathrm{D}}},\, \delta_{x_t}\right)}_{\text{cross-entropy on ground truth}} \\
&\quad +\;
\lambda \tau^{2}\,\mathbb{E}_{x\sim \mathcal{D}}\,\sum_{t=1}^{|x|}
    \underbrace{\mathrm{KL}\!\left(P_{M_T} \;\|\; P_{M_{\mathrm{D}}}\right)}_{\text{knowledge distillation}},
\end{split}
\]
where, $\delta_{x_t}$ is the one-hot distribution on the observed token $x_t$, $\mathrm{CE}$ denotes the cross-entropy, and $\lambda\in[0,1]$ balances the two terms. \(\theta\) are the parameters of \distilled{}. Here \(P_{M}\) is used as short hand for \(P_{M}(\cdot \mid x_{<t})\) for each model. Each probability distribution used in the \(\mathrm{KL}\) term is obtained from the logits $\mathbf{z}\in\mathbb{R}^{|\mathcal{V}|}$ with $P_M=\mathrm{softmax}(\frac{\mathbf{z}}{\tau})$ where \(\tau\) is the softmax temperature. The factor of \(\tau^2\) is standard in knowledge distillation \cite{hinton2015distilling} and is used to ensure the gradients from the cross-entropy and KL-divergence term are scaled similarly. We use \(\lambda=0.7\) and \(\tau=2\), which were obtained after a hyperparameter study (Appendix \ref{app:hyperparameters}). See Appendix \ref{app:distilled_reference_training} for more training details.

Knowledge distillation in this case only requires access to the logits of the target model. In general, the vocabularies of \target{} and \distilled{} may be different. When the vocabularies of the two models are different, it is necessary to align tokens when computing the KL divergence. This is a well-studied problem in the literature \cite{chen-etal-2025-enhancing-cross, boizard2024towards}. We do not perform an explicit alignment, as the target and reference models in our study either share the same tokenizer or a similar one. 




\subsection{Hypothesis Testing via Rank-Correlation Comparisons}

Define the Spearman rank correlation between the \minkpp{} scores ouput by  models $M_A$ and $M_B$ over \data{} as
\begin{multline*}
\rho(M_A,M_B;\mathcal{D}) \;\equiv\;\\ \mathrm{corr}\big(\mathrm{rank}(\{f_{\text{Min-K\%++}}(x^{(i)}; M_A)\}_{i=1}^n),\;\\\mathrm{rank}(\{f_{\text{Min-K\%++}}(x^{(i)}; M_B)\}_{i=1}^n)\big)
\end{multline*}
where \(\mathrm{rank}(.)\) maps scores to ranks within \data{} and \(\mathrm{corr}(.)\) is the Pearson correlation coefficient operator. We use the short-hand notation \(\rho_{AB}\) for this.
\paragraph{Test statistic.}
Our method compares the target’s rank agreement with the reference versus with the distilled reference. Let
\[
\widehat{\Delta}\;\equiv\; \rho\!\left(M_R,M_{T};\mathcal{D}\right)\;-\;\rho\!\left(M_D,M_T;\mathcal{D}\right).
\]

\[
\widehat{\Delta}\;\equiv\; \rho_{RT} - \rho_{FT}
\]


\paragraph{Bootstrap inference.}
As we are comparing two dependent correlations, we use a bootstrap percentile to compute a p-value \cite{wilcox2016comparing, dhingra2019handling, wilcox2017modern}. We construct a test on $\widehat{\Delta}$ using nonparametric bootstrap over sequences:
\begin{enumerate}[leftmargin=1.5em,itemsep=0.25em]
\item For $b=1,\dots,B$ (with $B{=}10{,}000$), draw a bootstrap resample $\mathcal{D}^{(b)}$ of size $n$ by sampling sequences from \data{} with replacement.
\item Compute $\rho^{(b)}_{DT}=\rho(M_D,M_{T};\mathcal{D}^{(b)})$ and $\rho^{(b)}_{RT}=\rho(M_R,M_T;\mathcal{D}^{(b)})$, and set
\(
\Delta^{(b)}=\rho^{(b)}_{RT}-\rho^{(b)}_{DT}.
\)
\end{enumerate}
$\{\Delta^{(b)}\}_{b=1}^B$ is known to approximate the sampling distribution of $\widehat{\Delta}$ \cite{efron1992bootstrap}.

\paragraph{Test for evidence of \emph{non}-training on \data{} (non-membership).}
The one-sided hypothesis test on the statistic $\widehat{\Delta}$ is as below where \(\Delta\) is the true value estimated by \(\widehat{\Delta}\):
\[
\begin{aligned}
&H_{0}^{\text{non}}:\; \Delta \leq 0
\quad\text{vs.}\quad
H_{1}^{\text{non}}:\; \Delta > 0.
\end{aligned}
\]

\[
\begin{aligned}
&H_{0}:\; \Delta \leq 0
\quad\text{vs.}\quad
H_{1}:\; \Delta > 0.
\end{aligned}
\]
The corresponding one-sided $p$-value is
\[
p \;=\; \frac{1+\#\{b:\,\Delta^{(b)}\le 0\}}{B+1},
\]
with rejection of $H_{0}^{\text{non}}$ at level $\alpha$ interpreted as evidence that $M_T$ \emph{was not} trained on $D$. We use a standard value of \(\alpha=0.05\).

The addition of one to the numerator and denominator is a finite sample correction which ensures that the p-value is always greater than 0 \cite{phipson2016permutation, ojala2010permutation}.

\section{Results}




\begin{table*}[htbp]
\centering
\caption{p-values for non-member detection. None of the datasets below were used while training the vanilla Pythia models, but were used to train Pythia-410m-CPT. Pythia-410m is used as the reference model for all cases except Pythia-410m and Pythia-410m-CPT where Pythia-1b was the reference model.}
\label{tab:non_member_detection}
\begin{tabular}{lcccccccc}
\toprule
Target Model & PubMed & HN & ArXiv & CC & Ubuntu & Freelaw & Enron & Reddit \\
\midrule
Pythia-410m-CPT & 1.000 & 0.354 & 0.113 & 0.302 & 0.162 & 0.982 & 0.128 & 0.817 \\
Pythia-410m & 1.0e-4 & 1.0e-4 & 1.0e-4 & 1.0e-4 & 1.0e-4 & 1.0e-4 & 2.0e-4 & 1.0e-4 \\
Pythia-1b & 1.0e-4 & 1.0e-4 & 1.0e-4 & 1.0e-4 & 1.0e-4 & 1.0e-4 & 1.0e-4 & 1.0e-4 \\
Pythia-2.8b & 1.0e-4 & 1.0e-4 & 1.0e-4 & 1.0e-4 & 1.0e-4 & 1.0e-4 & 1.0e-4 & 1.0e-4 \\
Pythia-6.9b & 1.0e-4 & 1.0e-4 & 1.0e-4 & 1.0e-4 & 1.0e-4 & 1.0e-4 & 2.0e-4 & 1.0e-4 \\
\bottomrule
\end{tabular}
\end{table*}


\begin{table*}[t]
    \centering
    \caption{p-values computed using \method{} on Pythia-410m-CPT using a split of each of the below datasets that was held out during continued pre-training}
    \label{tab:pvalues_fp}
    \begin{tabular}{lcccccccc}
        \toprule
        & PubMed & HN & ArXiV & CC & Reddit & Ubuntu & Freelaw & Enron \\
        \midrule
        p-value & 0.84 & 0.01 & 1.0e-4 & 0.02 & 0.03 & 1.0e-4 & 1.0e-4 & 1.0e-4 \\
        \bottomrule
    \end{tabular}
\end{table*}

\begin{table*}[htbp]
\centering
\caption{p-values for dataset level membership detection}
\label{tab:member_detection}
\resizebox{\textwidth}{!}{%
\begin{tabular}{lcccccccccc}
\toprule
Method & \multicolumn{2}{c}{PubMed} & \multicolumn{2}{c}{HN} & \multicolumn{2}{c}{ArXiv} & \multicolumn{2}{c}{CommonCrawl} & \multicolumn{2}{c}{Reddit} \\
 & 410m & 410m-CPT & 410m & 410m-CPT & 410m & 410m-CPT & 410m & 410m-CPT & 410m & 410m-CPT \\
\midrule
LLM-DI & 0.604 & 0.775 & 0.395 & \textbf{0.022} & 0.781 & 0.083 & 0.642 & \textbf{0.045} & 0.437 & 0.064 \\
PaCoST & \textbf{3.0e-13} & \textbf{0.045} & 0.683 & 0.999 & 0.158 & 0.122 & \textbf{0.002} & \textbf{0.009} & \textbf{0.001} & \textbf{4.0e-9} \\
\bottomrule
\end{tabular}
}%
\end{table*}

\textbf{Reference models}: We use the Pythia model series \cite{biderman2023pythia} and OLMo-1b \cite{groeneveld-etal-2024-olmo} model as reference models, as the datasets used to train these models, as well as their validation datasets, are publicly available, allowing us to precisely identify data that was excluded from training. All Pythia reference models are the `deduped' checkpoints trained on a deduplicated version of the Pile.

\noindent\textbf{Datasets}: We use datasets that have not previously been used to train our target model or reference model of interest, by using their validation sets. This ensures that any observed differences in rank correlation are attributable to training exposure rather than incidental overlap in pretraining corpora.

\begin{enumerate}[leftmargin=*]

\item \textbf{The Pile}: We use the deduplicated subsets of the Pile \citep{gao2020pile} from the domains of Wikipedia (Wiki), and Pubmed Central abstracts (PubMed), ArXiV, Common Crawl (CC) released by \citet{duan2024membership}. These datasets were deduplicated to be especially challenging for MIA methods. We also use the Ubuntu-IRC chats, Enron emails, and Freelaw datasets that were also held out from training the Pythia models and released by \citet{mainidi2024}. The Pythia models are used as reference models for all Pile datasets.

\item \textbf{Dolma}: We use the Reddit held-out subset of Dolma obtained from Paloma \citep{soldaini2024dolma, magnusson2024paloma}. This was further processed to ensure that all documents had a timestamp after the cut-off date of the Pythia models to ensure they were not seen during their training. As this dataset was known to be held out from training the OLMo models, we use the OLMo-1b as the reference model for the Reddit dataset. This allows us to test the effect of having a different architecture for the target and reference model. The OLMo and Pythia model series share a similar tokenizer with the OLMo tokenizer, having some additional special tokens for masking personally identifiable information.

See Appendix \ref{app:datasets} for additional details on datasets.
\end{enumerate}






We continue pretraining the Pythia-410m model, which we call Pythia-410m-CPT, to use as a target model. Each dataset from the sources above was divided into two parts, one of which was used for training and one was held out from training. In addition to the datasets above, we sample 6 billion tokens from the Common Pile dataset \citep{kandpal2025common} for pretraining (See Appendix \ref{app:continued_pretraining}). Each dataset used for training constitutes less than 0.001 \% of the pretraining corpus. 

\begin{table*}[htbp]
\centering
\caption{p-values for different reference models used for non-membership detection on Pythia-410m and Pythia-410m-CPT}
\label{tab:scoring_model_ablation}
\resizebox{\textwidth}{!}{%
\begin{tabular}{lcccccccc}
\toprule
Reference Model & \multicolumn{2}{c}{PubMed} & \multicolumn{2}{c}{HN} & \multicolumn{2}{c}{ArXiv} & \multicolumn{2}{c}{CommonCrawl} \\
 & 410m & 410m-CPT & 410m & 410m-CPT & 410m & 410m-CPT & 410m & 410m-CPT \\
\midrule
Pythia-70m & 0.080 & 0.858 & \textbf{1.0e-4} & \textbf{0.002} & \textbf{2.0e-4} & \textbf{0.030} & \textbf{0.043} & \textbf{0.030} \\
Pythia-160m & \textbf{1.0e-4} & 0.461 & \textbf{2.0e-4} & \textbf{0.017} & \textbf{1.0e-4} & \textbf{0.012} & \textbf{1.0e-4} & \textbf{4.0e-4} \\
Pythia-1b & \textbf{1.0e-4} & 1.000 & \textbf{1.0e-4} & 0.354 & \textbf{1.0e-4} & 0.113 & \textbf{1.0e-4} & 0.302 \\
\bottomrule
\end{tabular}
}%
\end{table*}

\subsection{Non-membership detection}

From the results of Table \ref{tab:non_member_detection}, we see that \method{} is able to detect non-members for all datasets correctly with a threshold of \(p<0.05\). When we test \method{} using Pythia-410-CPT, all p-values are well above our threshold, indicating that \method{} did not falsely rule out any datasets as being non-members. Crucially, these results also hold true for the Reddit dataset, which used a reference model (OLMo-1b) of a different architecture than the target model. The values of \(\rho_{RT}\), \(\rho_{DT}\), and the confidence interval for the difference in rank correlation corresponding to Table \ref{tab:non_member_detection} are provided in Appendix \ref{app:correlations_raw}. To further evaluate the robustness of \method{}, we compute p-values using splits of each of our datasets that were held out from training Pythia-410m-CPT (Table \ref{tab:pvalues_fp}). In all cases except one, \method{} is able to detect non-membership correctly. In the case of PubMed, the test is inconclusive, which is one possible outcome of the test. The lowest value observed is \(10^{-4}\) as 10,000 bootstrap samples are drawn, and this is thus the lowest p-value that can be obtained due to the finite sample correction in the computation of the p-value. See Appendix \ref{app:additional_results} for additional results.

We also study the performance of existing dataset-level membership detection approaches, LLM-DI and PaCoST \cite{zhang2024pacost}, at the task of dataset-level membership detection using Pythia-410m and Pythia-410m-CPT (See Appendix \ref{app:tdd_baselines} for more details). When using a strict p-value threshold of 0.05, LLM-DI can detect only two out of five datasets correctly (Table \ref{tab:member_detection}). LLM-DI has high p-values for three datasets included in the training, indicating that the test failed to conclusively determine whether these datasets were included in the training. Thus, non-membership detection is not simply the converse of detecting membership. A high p-value is not a guarantee of non-membership. In the case of PaCoST, we find that regardless of whether the dataset was included in training, membership was either indicated for both Pythia-410m and Pythia-410m-CPT or could not be concluded for either. Thus, the test does not provide a meaningful differentiating signal. PaCoST works by computing the confidence of a given document against its paraphrases, assuming that documents included in training will have higher confidence relative to paraphrases compared to documents not included in training. As pointed out in \citet{rastogi2025stampcontentprovingdataset}, the paraphrasing itself introduces a distribution shift, causing the test to detect lexical shifts instead of training signal.

\subsection{Ablations}

\paragraph{Ablating over the reference models.}

We vary the reference model used with \method{} for performing non-member detection with Pythia-410m and Pythia-410m-CPT (Table \ref{tab:scoring_model_ablation}). 
Across all datasets, Pythia-1b provides the most consistent signal.
Smaller models have lower representational capacity and therefore exhibit weaker alignment in rank correlations with the target model.
These results highlight that stronger and more stable correlations are obtained when the reference model has sufficient scale and capacity.


\paragraph{Varying the size of the dataset.}

\begin{figure}
    \centering
    \includegraphics[width=0.95\linewidth]{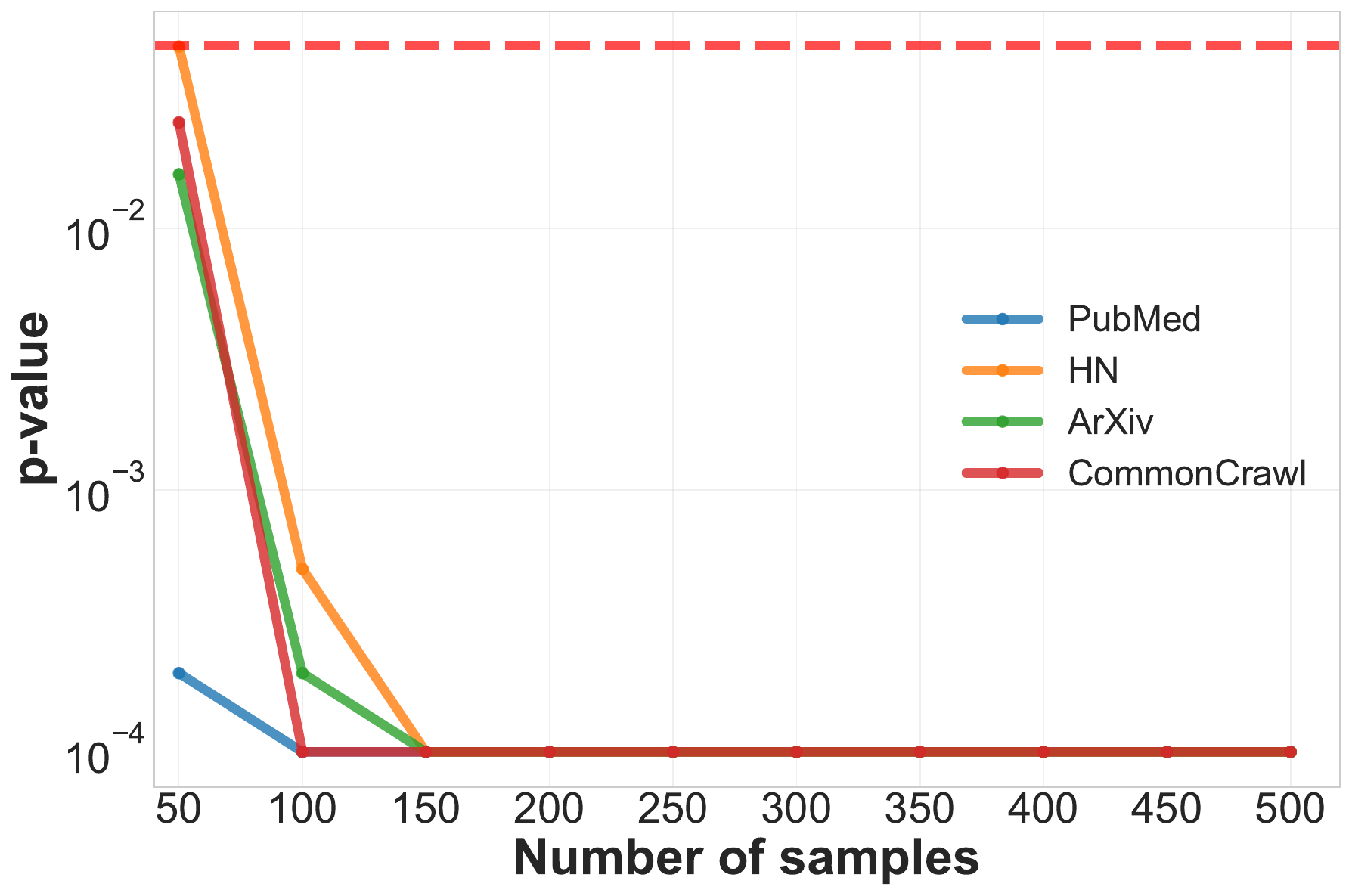}
    \caption{p-value for non-member detection when different number of samples are used}
    \label{fig:dataset_size_ablation}
\end{figure}

In Figure \ref{fig:dataset_size_ablation}, we see that as the size of the dataset is reduced, the p-value for non-member detection stays stable until fewer than 150 documents are used, thus making this very practical for real-world audit scenarios.


\subsection{Effect of \minkpp{}}

We analyze the behavior of \minkpp{} score in this section by taking the ArXiV dataset as an example. See Figure \ref{fig:min_k_ablation_app} and Figure \ref{fig:mink_rank_app} in the Appendix for plots corresponding to other datasets.

\paragraph{Changing the value of \(K\).} 
As the value of \(K\) increases, we observe that the correlation difference between Pythia-410m and Pythia-410m-CPT against a reference model drops.
This trend indicates that the strongest training signal is concentrated in the smallest \(K\%\) of each document’s normalized token log-probabilities.
Larger values of \(K\) include more predictable tokens, diluting the influence of the high-surprisal tokens that are most affected by memorization.

\begin{figure}[!h]
    \centering
    \begin{subfigure}[b]{0.8\columnwidth}
        \centering
        \includegraphics[width=\textwidth]{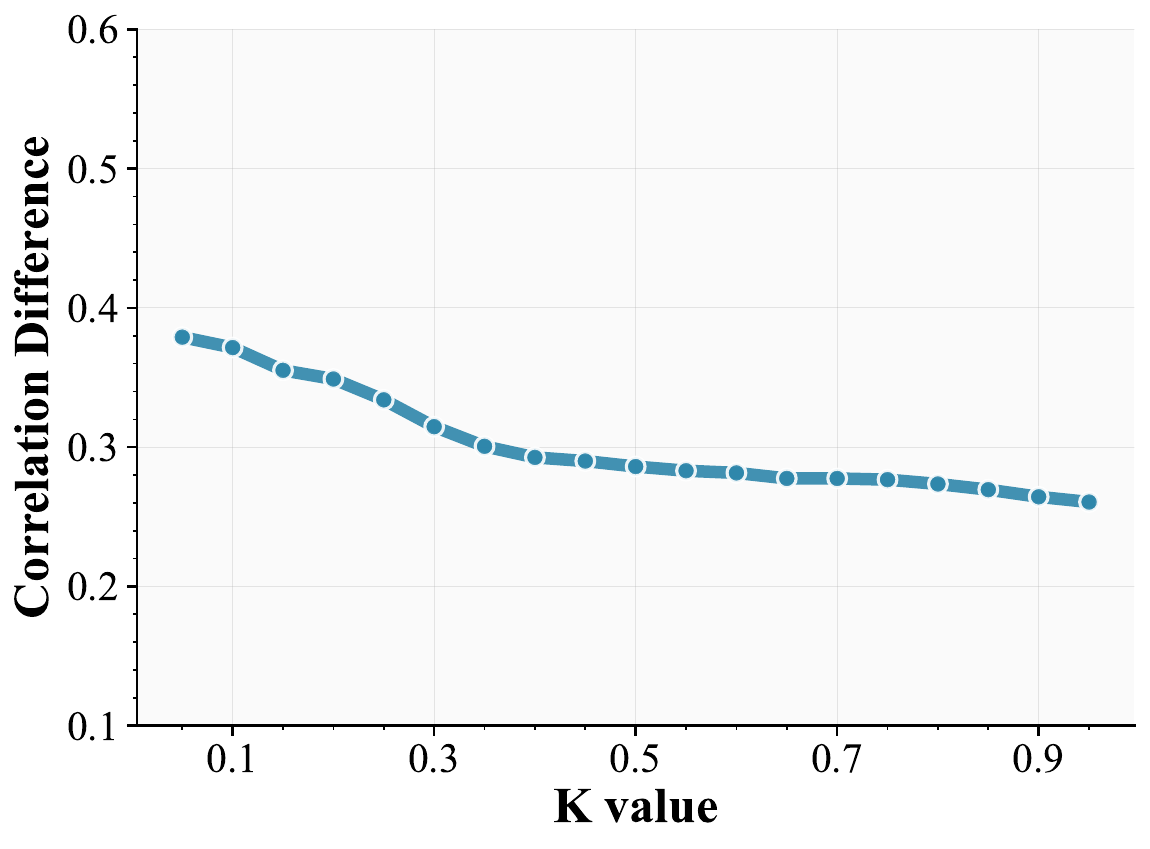}
        \caption{}
        \label{fig:subfig_a}
    \end{subfigure}
    \hfill
    \begin{subfigure}[b]{0.8\columnwidth}
        \centering
        \includegraphics[width=\textwidth]{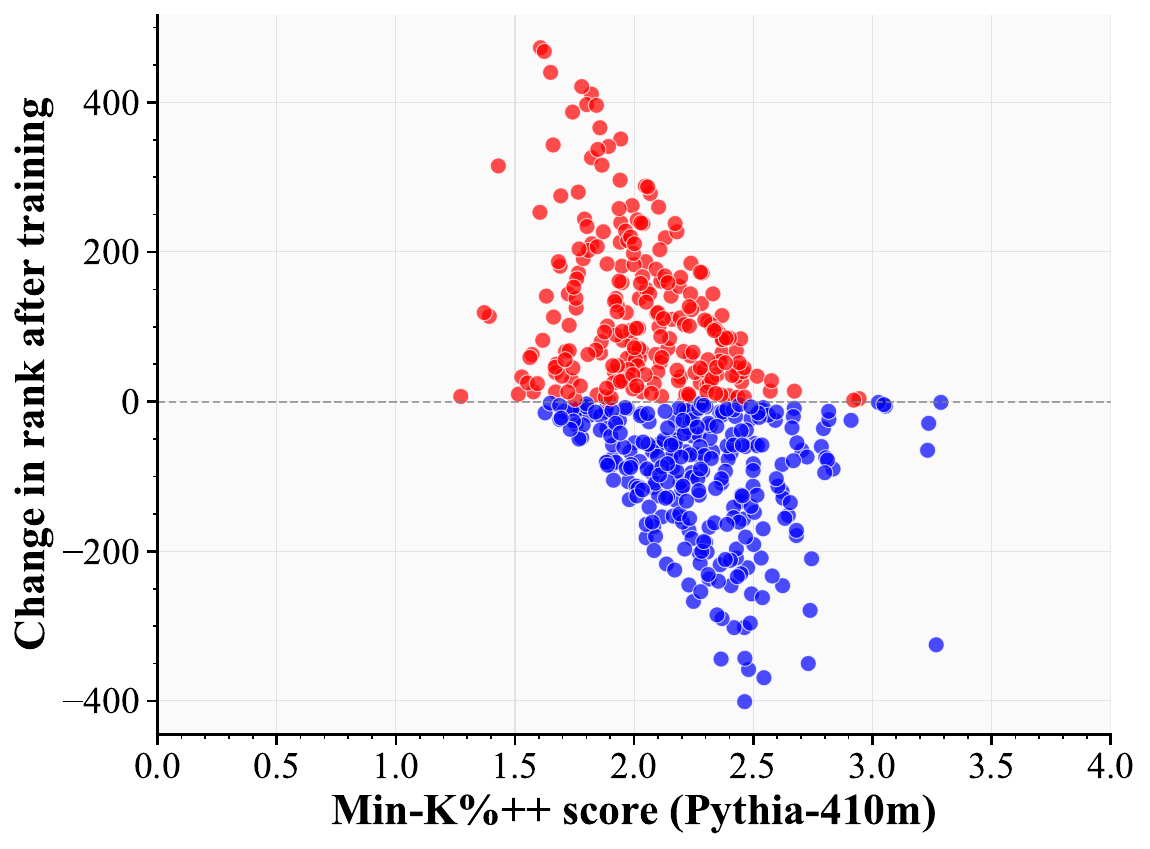}
        \caption{}
        \label{fig:subfig_b}
    \end{subfigure}
    
    \caption{a) Spearman rank correlation difference of Pythia-410m and Pythia-410m-CPT against Pythia-1b reference model at different values of \(K\) in \minkpp{} b) Change in rank of a document when \minkpp{} is computed using Pythia-410m-CPT relative to the rank of the Pythia-410m model.}
    \label{fig:minkpp_main}
\end{figure}


\paragraph{Studying change in rank with \minkpp{} value.}
We study the change in rank of \minkpp{} scores between Pythia-410m and Pythia-410m-CPT as a function of the \minkpp score of Pythia-410m (Figure \ref{fig:minkpp_main}b). The \minkpp{} score has been negated from the original definition \cite{zhang2025mink} so that higher \minkpp{} indicates higher average surprisal of the bottom \(K\) tokens for the model. Ranks are assigned in ascending order of \minkpp{} score. The results indicate that documents initially having a high \minkpp{} score drop in rank after training, while documents with low \minkpp{} scores go up in rank. 
This suggests that documents originally judged as “difficult” by the model become easier—showing reduced surprisal—once they are included in training. Thus, high-\minkpp{} documents exhibit the strongest memorization effect, supporting our use of \minkpp{} as a robust signal for detecting non-membership.

\section{Conclusions}

We presented \method{}, a novel method for showing with high confidence that a dataset has \emph{not} been used for training. While much past work has focused on detecting members, we flip the question to ask if non-member datasets can be detected with high confidence. We find that across eight datasets, \method{} is indeed able to detect non-members while not falsely ruling out datasets that were used during training. \method{} does not require access to a reference dataset, a common requirement for many other methods, and hence makes it a practical tool for compliance audits of LLMs.

\section*{Limitations}

\method{} relies on identifying when the dataset was released and for certain types of data, this may not be straightforward to establish. \method{} will be applicable for newer data after open-source LLMs became widely available. Our continued pre-training for Pythia-410m-CPT serves as an approximation to full pre-training of a model, and longer training runs may affect our results. \method{} also assumes that an entire dataset was either included or not included in training and cannot handle cases when a dataset was only partially included.

\section*{Disclaimer}
This paper was prepared for information purposes by the Artificial Intelligence Research group of JPMorgan Chase \& Co and its affiliates (“JP Morgan”) and is not a product of the Research Department of JP Morgan. J.P. Morgan makes no representation and warranty whatsoever and disclaims all liability for the completeness, accuracy, or reliability of the information contained herein. This document is not intended as investment research or investment advice, or a recommendation, offer, or solicitation for the purchase or sale of any security, financial instrument, financial product, or service, or to be used in any way for evaluating the merits of participating in any transaction, and shall not constitute a solicitation under any jurisdiction or to any person if such solicitation under such jurisdiction or to such person would be unlawful.

\bibliography{custom}

\appendix

\section*{Appendix}

\section{Datasets: Additional details}
\label{app:datasets}

We pick the datasets from \citet{duan2024membership} that were deduplicated against the Pile training data with a 13-gram Bloom filter and a threshold of 80 \% overlap. This means that the dataset was constructed from the held-out set of the Pile to avoid any document with up to a 13-gram overlap with any document included in the training set. Each training document was limited to between 100 and 200 words and at most 512 tokens. We control for the length as longer sequences make the document more detectable to MIA methods \citep{puerto-etal-2025-scaling}. 

\noindent\textbf{Reddit}: We use the Reddit subset of Dolma available in the Paloma dataset \citep{magnusson2024paloma}. To further deduplicate it, we compare the `ID' field of this data against the entire Dolma dataset and remove all documents from Paloma whose `ID' was found in Dolma. This left us with 1300 documents, which we truncated as above to 512 tokens with between 100 and 200 words. We further split this into 650 documents used for training and 650 documents used as held-out data.
The number of documents and tokens in each dataset is shown in Table \ref{tab:dataset_sizes}. All datasets and models used in our work are released under permissive open source licenses.

\begin{table}[htbp]
\centering
\caption{Dataset statistics}
\label{tab:dataset_sizes}
\begin{tabular}{lrr}
\toprule
Dataset & Num. Documents & Tokens \\
\midrule
PubMed & 500 & 154,387 \\
HN & 500 & 160,637 \\
ArXiv & 500 & 162,267 \\
CommonCrawl & 500 & 129,029 \\
Reddit & 650 & 125,789 \\
Ubuntu & 688 & 294,341 \\
Freelaw & 990 & 278,770 \\
Enron & 359 & 118,004 \\
\bottomrule
\end{tabular}
\end{table}

\section{Continued Pretraining of Pythia-410m}
\label{app:continued_pretraining}

In addition to the watermarked datasets, we sample documents from the Common Pile dataset for continued pre-training of Pythia-410m. Specifically, we select documents from the USPTO, USGPO, ArXiv, LibreText, and Doab domains of the Common Pile \citep{kandpal2025common}. Each document is split into sequences of 512 tokens and used for training after randomizing the order. To avoid overlap with data seen during initial pre-training, we only sample documents published after January 2021, which is the cut-off date of the Pile dataset.

We use the AdamW optimizer with a learning rate of \(10^{-4}\), \((\beta_1, \beta_2) = (0.99, 0.999)\), cosine decay, and a batch size of 40. A warmup of 0.5\% of training tokens was used with no weight decay. We train all our models on an L40S Tensor Core GPU. We utilize the Transformers library(version 4.43) and a random seed of 1234 for all our experiments \cite{wolf2019huggingface}.

\section{Training distilled reference model}
\label{app:distilled_reference_training}
For each distilled reference model, we train on the corresponding dataset for a single epoch with a learning rate of \(5 \times10^{-5}\) and a warmup of 5\% of the total steps. We use a batch size of 4 and 4 gradient accumulation steps. 

\section{Training data detection baselines}
\label{app:tdd_baselines}

\begin{enumerate}
    
    \item \textbf{LLM-DI}: LLM-DI \citep{mainidi2024} utilizes multiple MIA scores as features to train a classifier for detecting training data. To ascertain whether a model has been trained on a dataset, the verifier must generate features from both the suspect dataset and the non-member dataset and train a classifier. Binary classification using any MIA score yields an ROC-AUC score close to 0.5 \cite{chen-etal-2025-statistical}. Thus distinguishing any given member document from non-member document is difficult for this classifier. However, when the score of this classifier is averaged over all documents in the dataset and compared against a non-member dataset, a strong signal is produced. To calibrate this signal, LLM-DI assumes access to a non-member dataset from the same distribution. When the average classifier score of a suspect dataset is lower than that of the non-member dataset, membership can be concluded. The threshold here is determined by hypothesis testing. Given the reference assumed, the equivalent way to infer non-membership is to assume access to a member dataset from the same distribution as the dataset of interest and to flip the formulation of LLM-DI. In general, this cannot be known for an arbitrary target model. Thus, addressing this problem requires introducing a different reference against which to calibrate non-membership, which is where \method{} comes in.
    
    In our setup, we used the splits of each dataset held out from training Pythia-410m-CPT (during continued pretraining) as the non-member documents.
    The implementation provided by the authors of LLM-DI was utilized in our study.

    \item \textbf{PaCoST}: PaCoST~\cite{zhang2024pacost} is a statistical method designed to detect QA benchmark contamination in LLMs. The approach works by first rephrasing each data input to create a counterpart with the same meaning and difficulty but different wording. Both the original and rephrased questions are then presented to the model, and the model is prompted to evaluate if the given QA pairs are correct or not based on `Yes' or `No' output.
    Next, the model’s confidence in its answers is estimated using the output probability of the token `Yes'.
    PaCoST compares the confidence scores for the original and rephrased questions using a paired samples t-test to determine if the model is significantly more confident on the original benchmark data. If the statistical test yields a p-value less than 0.05, it indicates likely contamination, meaning the model has probably seen the benchmark data during training. 
    PaCoST was designed to detect contamination for QA benchmarks. To use PaCoST for our datasets, we have modified the prompts for rephrasing and confidence estimation. The modified prompts are given below.

\end{enumerate}


\tcbset{
  mypromptbox/.style={
    colback=gray!15,
    colframe=gray!50!black,
    boxrule=0.8pt,
    arc=4pt,
    outer arc=4pt,
    boxsep=5pt,
    left=5pt,
    right=5pt,
    top=5pt,
    bottom=5pt,
    fonttitle=\bfseries,
    title=PaCoST Confidence Prompt,
    coltitle=white,
    sharp corners=south,
    width=0.44\textwidth, 
  }
}

\begin{flushright}
\begin{tcolorbox}[mypromptbox]
You are an expert in judging whether the text is correct. 
        You will be given a text. 
        Your job is to determine whether this text is correct. 
        You should only respond with Yes or No. 
        For example, given question "The value of 1+1 is 2.", the correct response should be "Yes".

\end{tcolorbox}
\end{flushright}

\tcbset{
  mypromptbox/.style={
    colback=gray!15,
    colframe=gray!50!black,
    boxrule=0.8pt,
    arc=4pt,
    outer arc=4pt,
    boxsep=5pt,
    left=5pt,
    right=5pt,
    top=5pt,
    bottom=5pt,
    fonttitle=\bfseries,
    title=PaCoST Rephrasing Prompt,
    coltitle=white,
    sharp corners=south,
    width=0.44\textwidth, 
  }
}
\begin{flushright}
\begin{tcolorbox}[mypromptbox]
You are provided with a text. 
Your task is to rephrase this text into another text with the same meaning.
When rephrasing the text, you must ensure that you follow the following rules: 
(1). You must ensure that you generate a rephrased text as your response. 
            (2). You must ensure that the rephrased text bears the same meaning with the original text. Do not miss any information. 
            (3). You must only generate a rephrased text. Any other information should not appear in your response. 
            (4). Do not output any explanation. 
            (5). Do not modify the numbers or quantities in the question. You should remain them unchanged. 
            Example 1: given text "1+1=2", one possible response should be: The value of 1+1 is 2. 
            Example 2: given text "Earth orbits around the Sun", one possible response should be: The Sun is the center of earth\'s orbit. 
            Example 3: given text "C is the third letter in English", one possible response should be: English\'s third letter is C. 
            Example 4: given text "A scientist is looking for something to watch faraway stars.", one possible response should be: A scientist would like use something to watch remote stars. 
            Example 5: given text "John has 8 cats and five dogs. Linda has 6 rabbits.", one possible response should be: John owns 8 cats and five dogs. Linda possesses 6 rabbits. 
 
\end{tcolorbox}
\end{flushright}

\section{Membership Inference}
\label{app:mia_methods}

Below, we summarize some key methods used in the literature.

\begin{itemize}
  \item \textbf{Perplexity/Loss} \citep{yeom2018privacy}. Uses the model’s loss (perplexity for language models) as the score: lower loss implies higher likelihood of membership. 

  \item \textbf{Min-K\%} \citep{shi2023detecting}. Averages the probabilities of the \emph{least likely} $K\%$ tokens in a sequence. 

  \item \textbf{Zlib Entropy} \citep{carlini2021extracting}. Computes a score as the ratio between model perplexity and the \texttt{zlib} compression size of the text. Smaller ratios indicate potential membership.

  \item \textbf{DC-PPD} \citep{zhang-etal-2024-pretraining}. Detects membership by measuring the divergence between the model’s token-probability distribution and the empirical token-frequency distribution of the text.
\end{itemize}


\section{Correlation trends}
\label{app:correlation_trends}
In Table \ref{tab:mia_scores_delta}, we compare the correlation difference using well-known MIA scores in addition to \minkpp{} to justify the choice of using \minkpp{}. The ranking is computed using each of the scores in the table for each dataset, where the reference model for computing correlations against was Pythia-1b. Observe that \minkpp{} was the score that consistently gave the largest differences, thus making it a suitable choice as a proxy for the training signal.


\begin{table*}[t]
    \centering
    \scriptsize
    \caption{Spearman rank correlation of Pythia-410m and Pythia-410m-CPT models against Pythia-1b reference model computed using different MIA scores}
    \label{tab:mia_scores_delta}
    \begin{tabular}{l*{15}{c}}
        \toprule
        & \multicolumn{3}{c}{\textbf{PubMed}} & \multicolumn{3}{c}{\textbf{HN}} & \multicolumn{3}{c}{\textbf{ArXiv}} & \multicolumn{3}{c}{\textbf{CommonCrawl}} & \multicolumn{3}{c}{\textbf{Wiki}} \\
        \cmidrule(lr){2-4} \cmidrule(lr){5-7} \cmidrule(lr){8-10} \cmidrule(lr){11-13} \cmidrule(lr){14-16}
        \textbf{Score} & 410m & CPT & $\Delta$ & 410m & CPT & $\Delta$ & 410m & CPT & $\Delta$ & 410m & CPT & $\Delta$ & 410m & CPT & $\Delta$ \\
        \midrule
        Loss & 0.99 & 0.92 & 0.07 & 0.98 & 0.83 & 0.15 & 0.99 & 0.84 & 0.15 & 0.98 & 0.72 & 0.26 & 0.98 & 0.55 & 0.43 \\
        zlib & 0.99 & 0.93 & 0.06 & 0.99 & 0.93 & 0.06 & 0.99 & 0.89 & 0.10 & 0.98 & 0.82 & 0.16 & 0.98 & 0.57 & 0.41 \\
        DC-PDD & 0.98 & 0.92 & 0.06 & 0.95 & 0.62 & 0.33 & 0.97 & 0.85 & 0.12 & 0.90 & 0.50 & \textbf{0.40} & 0.98 & 0.47 & 0.51 \\
        Min-K\% & 0.96 & 0.79 & 0.17 & 0.90 & 0.64 & 0.26 & 0.93 & 0.59 & 0.34 & 0.94 & 0.75 & 0.19 & 0.94 & 0.42 & 0.52 \\
        \minkpp{} & 0.84 & 0.33 & \textbf{0.51} & 0.75 & 0.41 & \textbf{0.34} & 0.82 & 0.45 & \textbf{0.37} & 0.79 & 0.51 & 0.28 & 0.84 & 0.25 & \textbf{0.59} \\
        \bottomrule
    \end{tabular}
\end{table*}



\section{Effect of Distillation Hyperparameters}
\label{app:hyperparameters}

\begin{figure}[t]
    \centering
    \begin{subfigure}[b]{0.78\columnwidth}
        \centering
        \includegraphics[width=\textwidth]{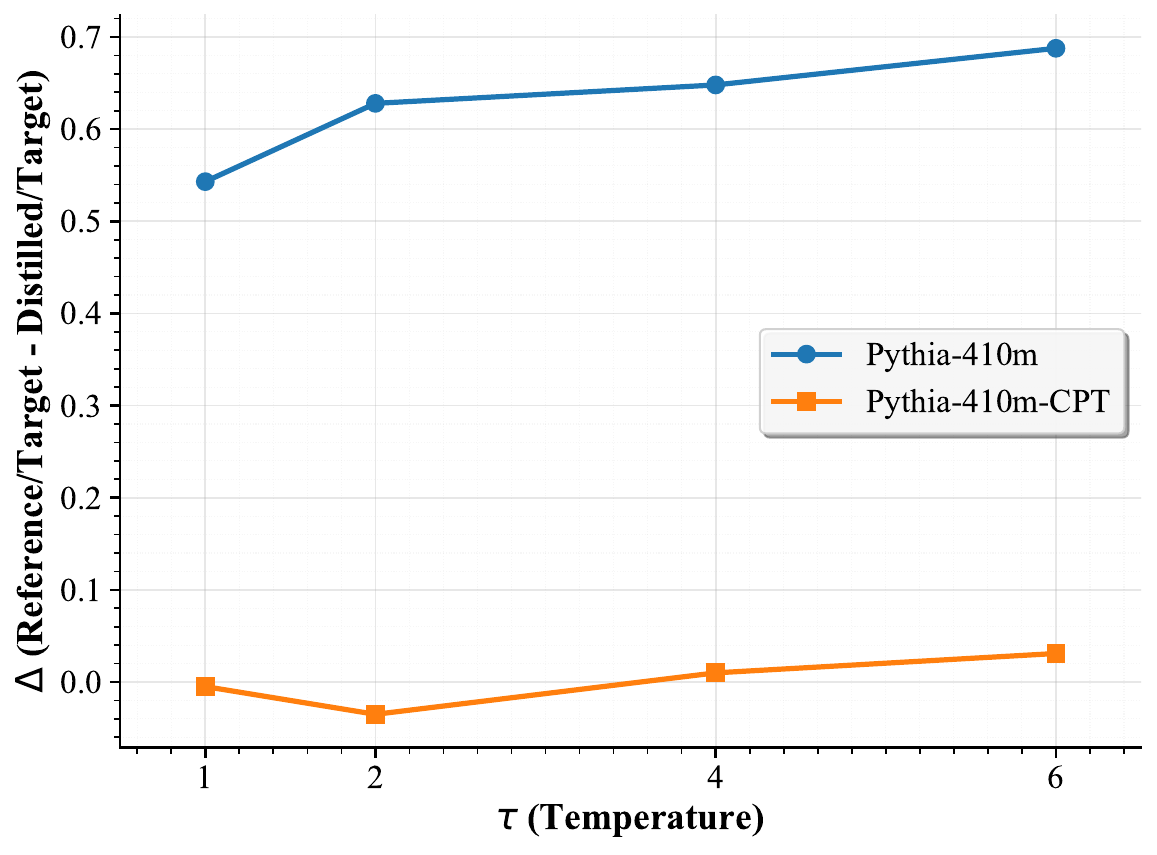}
        \caption{}
        \label{fig:subfig_a_hyperparam}
    \end{subfigure}
    \vspace{0.5em}
    \begin{subfigure}[b]{0.78\columnwidth}
        \centering
        \includegraphics[width=\textwidth]{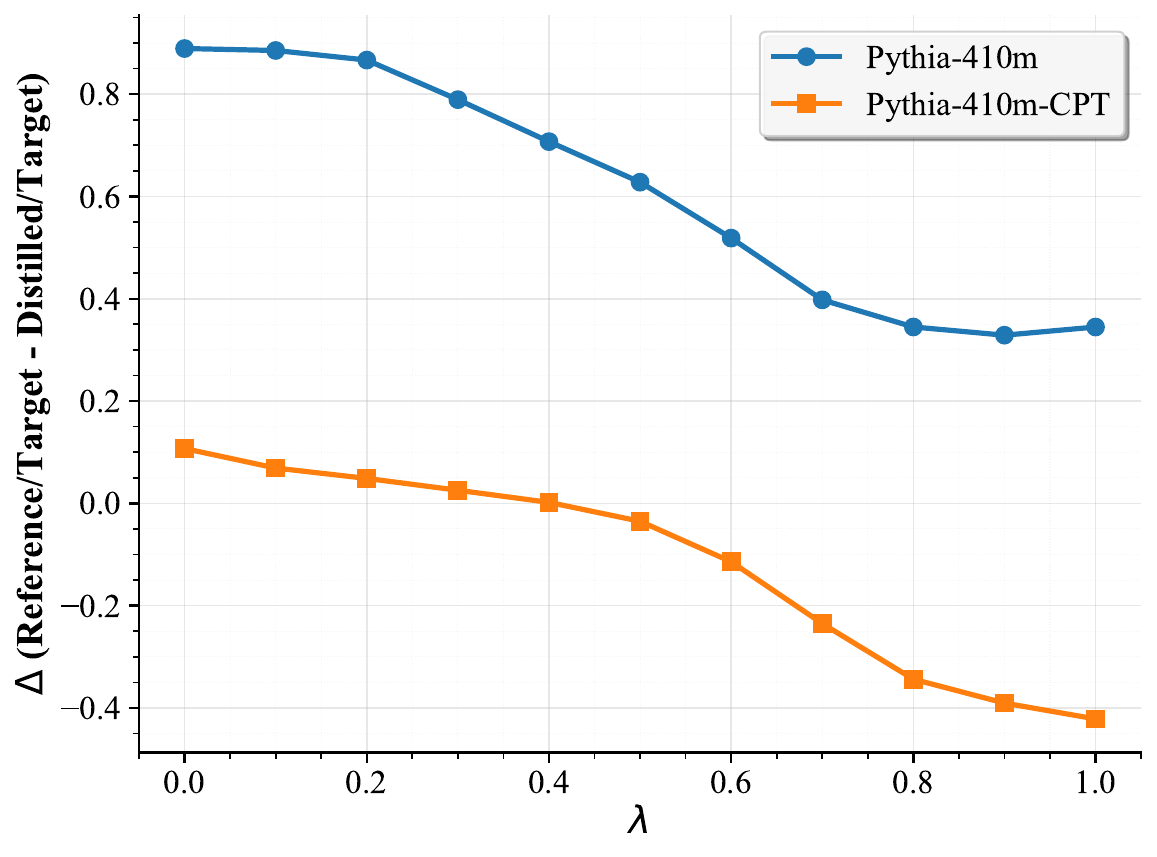}
        \caption{}
        \label{fig:subfig_b_hyperparam}
    \end{subfigure}
    \caption{Hyperparameter study to determine the value of a) \(\tau\) and b) \(\lambda\). Pythia-410m and Pythia-410m-CPT are the target models while Pythia-1b is the reference model.}
    \label{fig:hyperparameter_study}
\end{figure}

We study the effect of the distillation hyperparameters—the \emph{temperature} (\(\tau\)) and the \emph{distillation weight} (\(\lambda\))—on the correlation difference using the Wikipedia dataset (Figure~\ref{fig:hyperparameter_study}). The temperature \(\tau\) controls the smoothness of the target model’s probability distribution during distillation, while \(\lambda\) balances the contribution of the knowledge-distillation loss and the cross-entropy loss on the suspect dataset. The goal is to pick \(\lambda\) and \(\tau\) such that the correlation difference is positive for Pythia-410m (detecting true positives) and negative for Pythia 410m-CPT to avoid the possibility of false positives.

We first vary \(\tau\) over a grid while fixing \(\lambda = 0.5\). We identify \(\tau = 2\) as the optimal value, as it yields a negative correlation difference for Pythia-410m-CPT, allowing members to be ruled out with high confidence. With \(\tau\) fixed at this value, we next vary \(\lambda\) between \(0\) and \(1\). We observe that the correlation difference decreases as \(\lambda\) increases for both Pythia-410m and Pythia-410m-CPT. 
We pick the smallest \(\lambda\) that prevents false positives. At \(\lambda=0.6\) the correlation difference is -0.1, which still leaves open the possibility of false positives, hence we pick \(\lambda=0.7\) where the correlation difference is -0.25. We do not pick a higher \(\lambda\) because that would lead to lower correlation difference for detecting true positives. Accordingly, we select \(\lambda = 0.7\) and \(\tau = 2\) as our default hyperparameters. These hyperparameters were applied to all datasets and target models of different sizes in Table \ref{tab:non_member_detection}.

The sensitivity to hyperparameters can also be observed from Figure~\ref{fig:hyperparameter_study}. Detecting true positives works at all hyperparameter values, as the correlation difference is always much greater than 0 for Pythia-410m. To avoid false positives, however, it is necessary to pick $\lambda\ > 0.6$.

\section{Rank correlation results}
\label{app:correlations_raw}
Tables \ref{tab:non_member_correlation_group_1} and \ref{tab:non_member_correlation_group_2} show the raw rank correlations and the corresponding confidence intervals for the p-values in Table \ref{tab:non_member_detection}.

\begin{table*}[htbp]
\caption{Spearman rank correlations using for each target model and dataset. This is the raw data corresponding to the first four datasets in Table \ref{tab:non_member_detection}. \(\rho_{RT}\) is the Spearman rank correlation between the \minkpp{} scores from the reference and target model over the dataset, \(\rho_{DT}\) is the Spearman rank correlation between the \minkpp{} scores from the distilled reference and target model and \(\Delta_{95\% CI}\) is the 95 \% confidence interval for \(\rho_{RT}-\rho_{DT}\) obtained by bootstrapping.}
\label{tab:non_member_correlation_group_1}
\resizebox{\textwidth}{!}{%
\begin{tabular}{lcccccccccccc}
\toprule
Target Model & \multicolumn{3}{c}{PubMed} & \multicolumn{3}{c}{HN} & \multicolumn{3}{c}{ArXiv} & \multicolumn{3}{c}{CommonCrawl} \\
 & $\rho_{RT}$ & $\rho_{DT}$ & $\Delta_{95\% CI}$ & $\rho_{RT}$ & $\rho_{DT}$ & $\Delta_{95\% CI}$ & $\rho_{RT}$ & $\rho_{DT}$ & $\Delta_{95\% CI}$ & $\rho_{RT}$ & $\rho_{DT}$ & $\delta_{95\% CI}$ \\
\cmidrule(lr){2-4}
\cmidrule(lr){5-7}
\cmidrule(lr){8-10}
\cmidrule(lr){11-13}
Pythia-410m-CPT & 0.328 & 0.590 & [-0.371, -0.151] & 0.404 & 0.384 & [-0.080, 0.121] & 0.447 & 0.393 & [-0.034, 0.142] & 0.512 & 0.488 & [-0.068, 0.116] \\
Pythia-410m & 0.836 & 0.420 & [0.322, 0.511] & 0.752 & 0.467 & [0.207, 0.366] & 0.819 & 0.506 & [0.232, 0.394] & 0.791 & 0.446 & [0.261, 0.434] \\
Pythia-1b & 0.844 & 0.513 & [0.249, 0.416] & 0.722 & 0.512 & [0.127, 0.291] & 0.812 & 0.539 & [0.200, 0.348] & 0.840 & 0.482 & [0.282, 0.439] \\
Pythia-2.8b & 0.783 & 0.498 & [0.203, 0.369] & 0.661 & 0.307 & [0.262, 0.446] & 0.760 & 0.443 & [0.239, 0.396] & 0.807 & 0.361 & [0.356, 0.536] \\
Pythia-6.9b & 0.757 & 0.385 & [0.283, 0.463] & 0.651 & 0.321 & [0.240, 0.420] & 0.733 & 0.452 & [0.204, 0.358] & 0.769 & 0.302 & [0.371, 0.563] \\
\bottomrule
\end{tabular}
}%
\end{table*}

\begin{table*}[htbp]
\caption{Spearman rank correlations using for each target model and dataset. This is the raw data corresponding to the last four datasets in Table \ref{tab:non_member_detection}. \(\rho_{RT}\) is the Spearman rank correlation between the \minkpp{} scores from the reference and target model over the dataset, \(\rho_{DT}\) is the Spearman rank correlation between the \minkpp{} scores from the distilled reference and target model and \(\Delta_{95\% CI}\) is the 95 \% confidence interval for \(\rho_{RT}-\rho_{DT}\) obtained by bootstrapping.}
\label{tab:non_member_correlation_group_2}
\resizebox{\textwidth}{!}{%
\begin{tabular}{lcccccccccccc}
\toprule
Target Model & \multicolumn{3}{c}{Ubuntu} & \multicolumn{3}{c}{Freelaw} & \multicolumn{3}{c}{Enron} & \multicolumn{3}{c}{Reddit} \\
 & $\rho_{RT}$ & $\rho_{DT}$ & $\Delta_{95\% CI}$ & $\rho_{RT}$ & $\rho_{DT}$ & $\Delta_{95\% CI}$ & $\rho_{RT}$ & $\rho_{DT}$ & $\Delta_{95\% CI}$ & $\rho_{RT}$ & $\rho_{DT}$ & $\delta_{95\% CI}$ \\
\cmidrule(lr){2-4}
\cmidrule(lr){5-7}
\cmidrule(lr){8-10}
\cmidrule(lr){11-13}
Pythia-410m-CPT & 0.312 & 0.259 & [-0.054, 0.160] & 0.411 & 0.490 & [-0.151, -0.006] & 0.432 & 0.355 & [-0.055, 0.208] & 0.449 & 0.488 & [-0.123, 0.045] \\
Pythia-410m & 0.784 & 0.266 & [0.429, 0.605] & 0.826 & 0.591 & [0.184, 0.289] & 0.703 & 0.440 & [0.137, 0.392] & 0.672 & 0.080 & [0.503, 0.678] \\
Pythia-1b & 0.767 & 0.395 & [0.302, 0.441] & 0.825 & 0.498 & [0.277, 0.377] & 0.646 & 0.287 & [0.226, 0.496] & 0.721 & 0.070 & [0.560, 0.736] \\
Pythia-2.8b & 0.718 & 0.308 & [0.331, 0.489] & 0.753 & 0.398 & [0.298, 0.413] & 0.530 & 0.188 & [0.202, 0.479] & 0.763 & 0.052 & [0.626, 0.793] \\
Pythia-6.9b & 0.675 & 0.298 & [0.295, 0.459] & 0.708 & 0.359 & [0.289, 0.410] & 0.443 & 0.178 & [0.126, 0.405] & 0.774 & 0.038 & [0.650, 0.817] \\
\bottomrule
\end{tabular}
}%
\end{table*}

\section{Additional results}
\label{app:additional_results}

\subsection{Additional baseline}

While PRISM uses rank correlations, we test whether directly using the \minkpp{} scores can be used to infer non-membership. We use the \minkpp{} scores from the target model and the distilled reference model. As the distilled reference model is trained on the data set, its \minkpp{} scores are expected to be higher than those of the target model if the target model has not been trained on the data set. Formally, the null hypothesis is that the \minkpp{} scores under both models are equal and can be written as:


\begin{align}
H_0: & \; \mathbb{E}_{x}[f_{\text{Min-K\%++}}(x; M_T)] \nonumber \\
     & = \mathbb{E}_{x}[f_{\text{Min-K\%++}}(x; M_D)]
\end{align}

while the alternate hypothesis can be written as:

\begin{align}
H_1: & \; \mathbb{E}_{x}[f_{\text{Min-K\%++}}(x; M_T)] \nonumber \\
     & < \mathbb{E}_{x}[f_{\text{Min-K\%++}}(x; M_D)]
\end{align}

Note that establishing non-membership requires comparing against a model where the dataset is known to be a member, and thus, to establish non-membership of \data{} in \(M_T\) requires using \(M_D\) as the reference. To compute a p-value, we perform a paired t-test over each document in the data set.

The results (Table \ref{tab:non_member_detection_baseline}) show that non-member detection is inconsistent, as evidenced by HackerNews and Enron for Pythia-2.b and Pythia-6.9b, where non-membership fails to be detected. For Pythia-410m-CPT, in all cases, non-membership is falsely concluded. This shows that using \minkpp{} scores alone is not a reliable way to infer non-membership. As \minkpp{} values from different models are being compared, the difference in \minkpp{} values between them would arise not only due to a difference in membership of the dataset in training but also due to differences in the number of parameters and the architectures of the models.

\begin{table*}[htbp]
\caption{Non-Member detection baseline results using \minkpp{} values from target and distilled reference model.}
\label{tab:non_member_detection_baseline}
\resizebox{\textwidth}{!}{%
\begin{tabular}{lcccccccc}
\toprule
Target Model & PubMed & HN & ArXiv & CommonCrawl & Ubuntu & Freelaw & Enron & Reddit \\
\midrule
Pythia-410m-CPT & 2.6e-5 & 1.7e-24 & 3.7e-45 & 8.0e-8 & 1.2e-16 & 9.9e-34 & 3.6e-8 & 2.5e-8 \\
Pythia-410m & 0.18 & 9.6e-22 & 9.5e-11 & 2.8e-21 & 0.54 & 0.03 & 5.7e-17 & 2.5e-6 \\
Pythia-1b & 1.9e-11 & 2.2e-3 & 2.6e-47 & 7.0e-10 & 1.2e-18 & 3.7e-35 & 0.84 & 3.3e-6 \\
Pythia-2.8b & 7.8e-15 & 0.63 & 2.3e-8 & 0.06 & 0.04 & 1.6e-31 & 0.36 & 2.4e-6 \\
Pythia-6.9b & 4.8e-5 & 1.0 & 2.8e-38 & 0.01 & 2.7e-4 & 0.03 & 0.92 & 2.0e-6 \\
\bottomrule
\end{tabular}
}%
\end{table*}

\subsection{OLMo-1b as the target model}

We test \method{} using OLMo-1b as the target model on the Reddit dataset with various Pythia models as the reference models (Table \ref{tab:olmo_pvalues}). In all cases, non-membership was successfully detected. The Reddit dataset is non-member for all the Pythia models, as all documents in it have a release date after the knowledge cut-off of the Pythia models. 
\begin{table*}[htbp]
\centering
\caption{p-values for different Pythia reference models with OLMo-1b as the target model for the Reddit dataset.}
\label{tab:olmo_pvalues}
\begin{tabular}{lcccc}
\toprule
 & Pythia-70m & Pythia-160m & Pythia-410m & Pythia-1b \\
\midrule
$p$-value & 1.0e-4 & 1.0e-4 & 1.0e-4 & 1.0e-4 \\
\bottomrule
\end{tabular}
\end{table*}

\subsection{Rank correlation metric}

While we have used the Spearman correlation metric for computing rank correlations everywhere in our work, we find that computing the rank correlation with Kendall \(\tau\) would also work equally well and gives similar results for non-membership detection (Table \ref{tab:spearman_kendall_comparison}).

\begin{table*}[htbp]
\centering
\caption{Non-Member detection p-values using \method{} where the rank correlation is computed using Spearman \(\rho\) and Kendall \(\tau\)}
\label{tab:spearman_kendall_comparison}

\begin{tabular}{lccccc}
\toprule
Correlation Method & PubMed & HN & ArXiv & CommonCrawl & Reddit \\
\midrule
Spearman $\rho$ & 1.0e-4 & 1.0e-4 & 1.0e-4 & 1.0e-4 & 1.0e-4 \\
Kendall $\tau$ & 1.0e-4 & 1.0e-4 & 1.0e-4 & 1.0e-4 & 1.0e-4 \\
\bottomrule
\end{tabular}
\end{table*}

\begin{figure}[t]
    \centering
    \begin{subfigure}[b]{0.49\columnwidth}
        \centering
        \includegraphics[width=\textwidth]{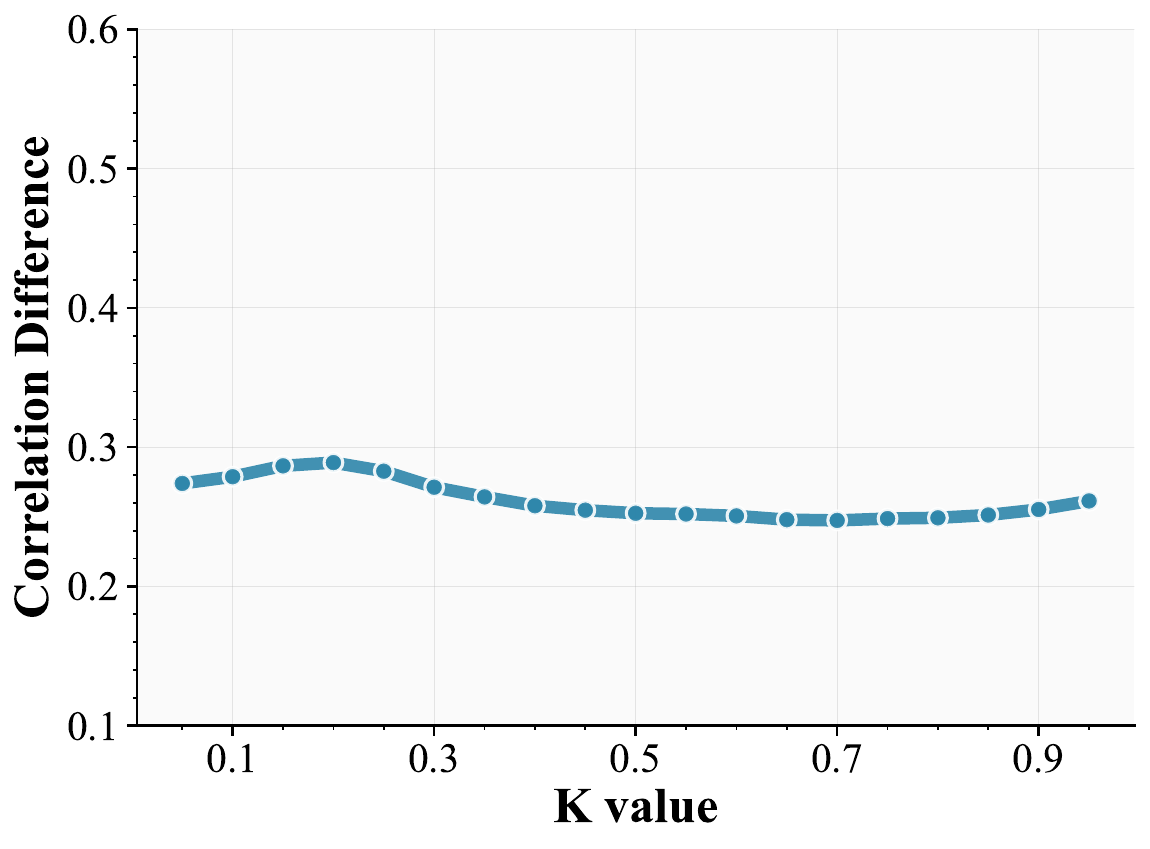}
        \caption{}
        \label{fig:subfig_a_correlation_difference}
    \end{subfigure}
    \hfill
    \begin{subfigure}[b]{0.49\columnwidth}
        \centering
        \includegraphics[width=\textwidth]{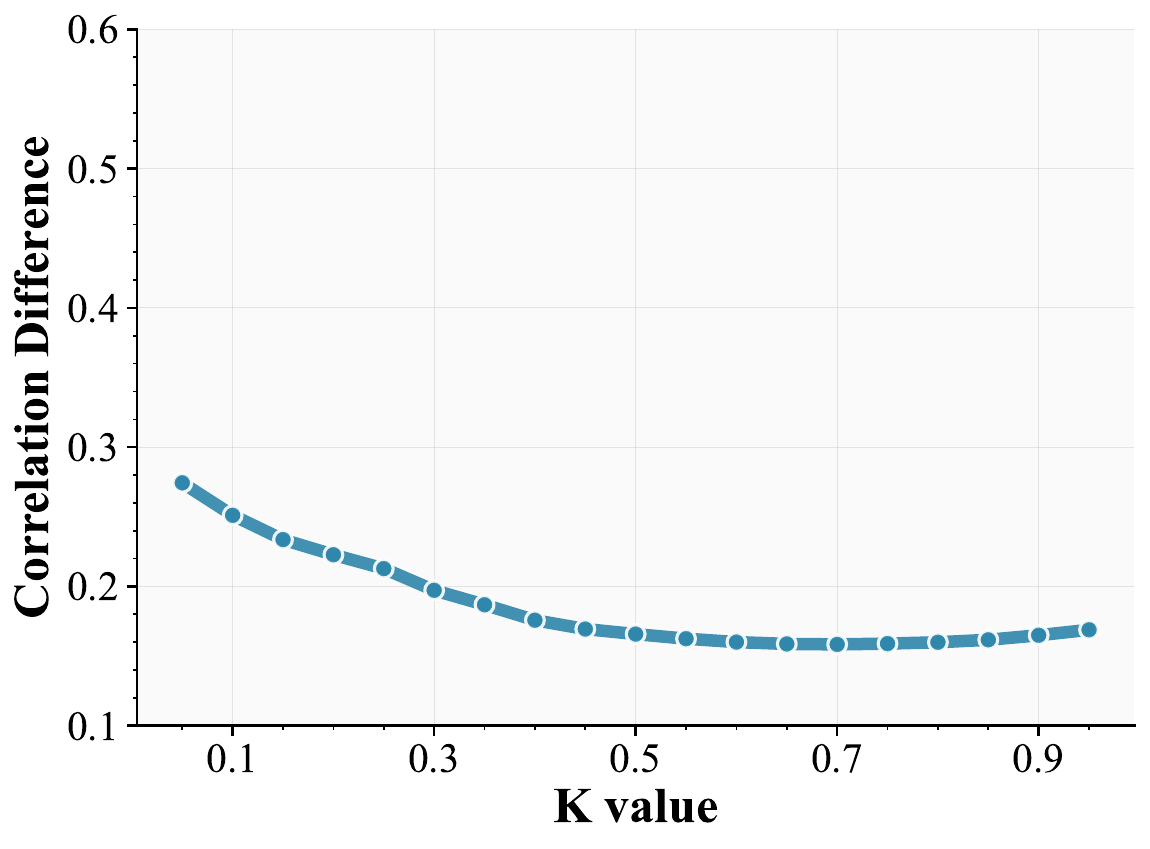}
        \caption{}
        \label{fig:subfig_b_correlation_difference}
    \end{subfigure}
    
    \vspace{0.5em}
    
    \begin{subfigure}[b]{0.49\columnwidth}
        \centering
        \includegraphics[width=\textwidth]{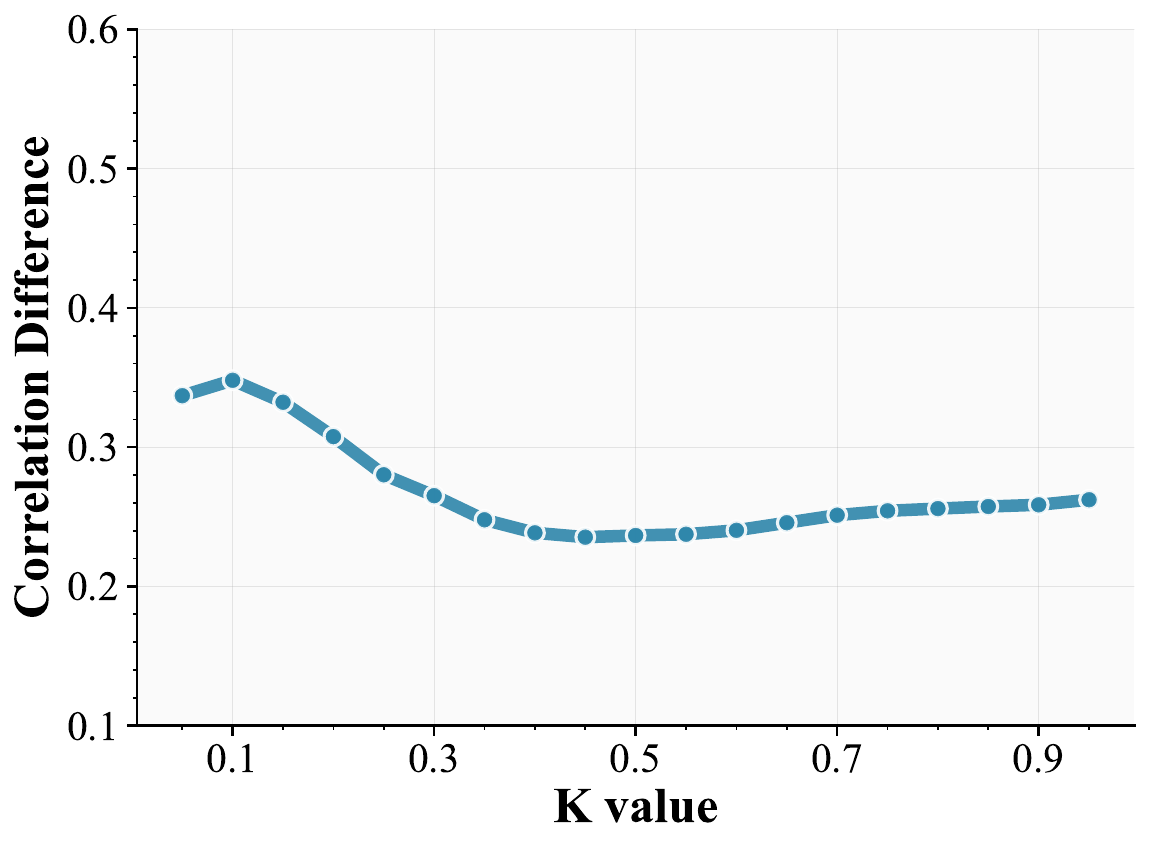}
        \caption{}
        \label{fig:subfig_c_correlation_difference}
    \end{subfigure}
    \hfill
    \begin{subfigure}[b]{0.49\columnwidth}
        \centering
        \includegraphics[width=\textwidth]{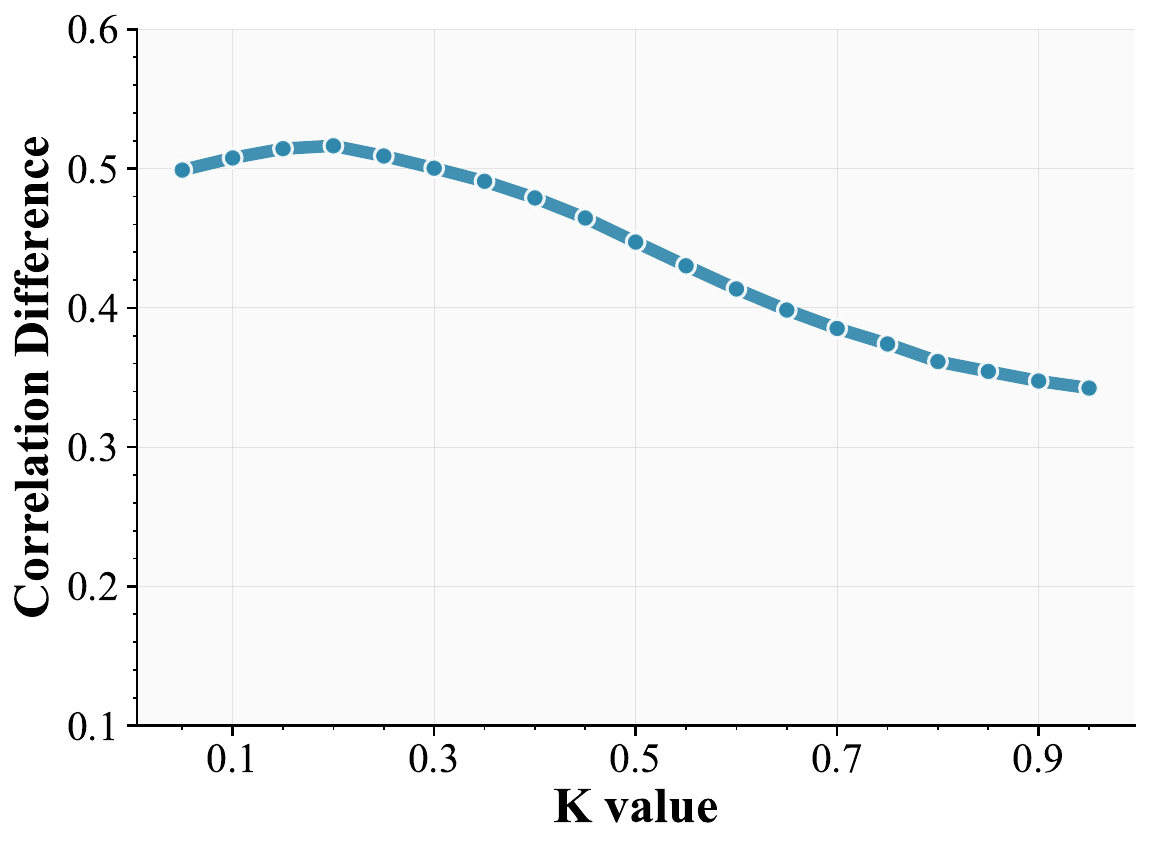}
        \caption{}
        \label{fig:subfig_d_correlation_difference}
    \end{subfigure}
    \caption{Spearman rank correlation difference of Pythia-410m and Pythia-410m-CPT against Pythia-1b reference model at different values of \(K\) in \minkpp{} for different datasets a) CommonCrawl b) Reddit c) HackerNews d) PubMed}
    \label{fig:min_k_ablation_app}
\end{figure}

\begin{figure}[t]
    \centering
    \begin{subfigure}[b]{0.49\columnwidth}
        \centering
        \includegraphics[width=\textwidth]{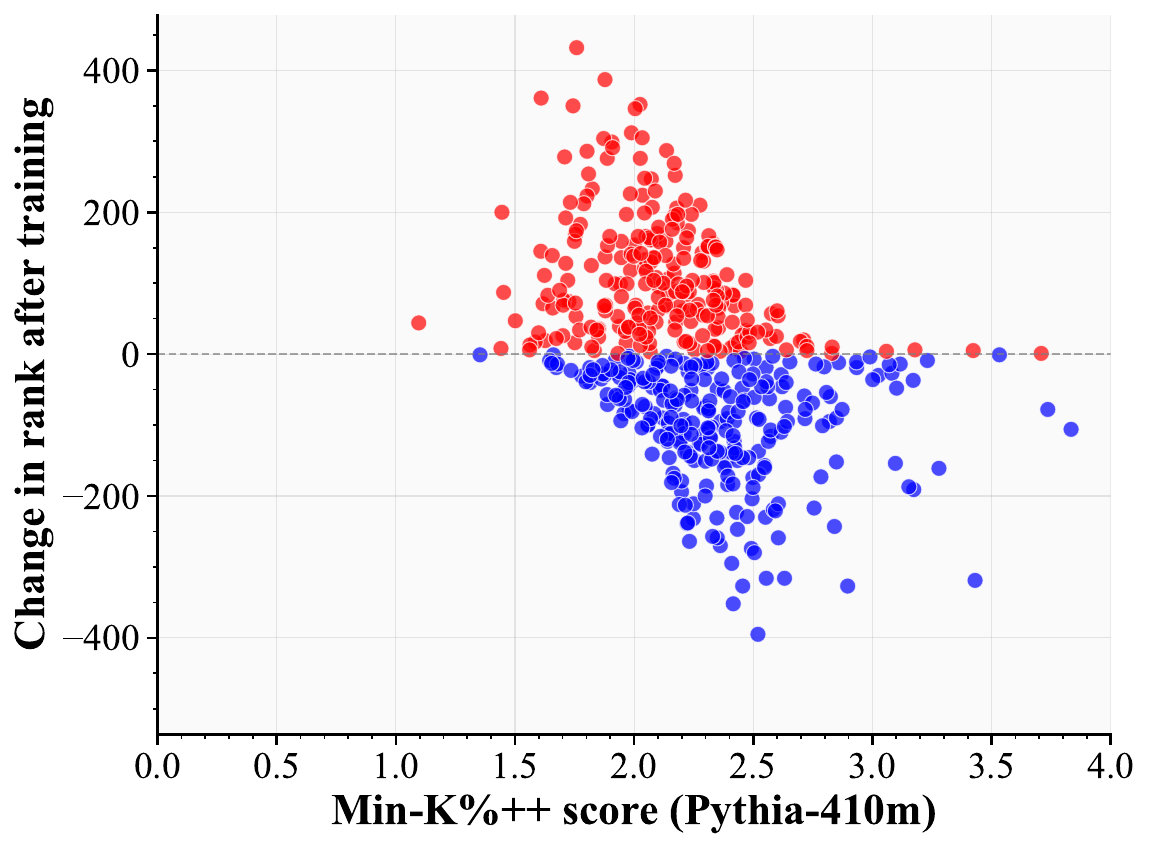}
        \caption{}
        \label{fig:subfig_a_rank_delta}
    \end{subfigure}
    \hfill
    \begin{subfigure}[b]{0.49\columnwidth}
        \centering
        \includegraphics[width=\textwidth]{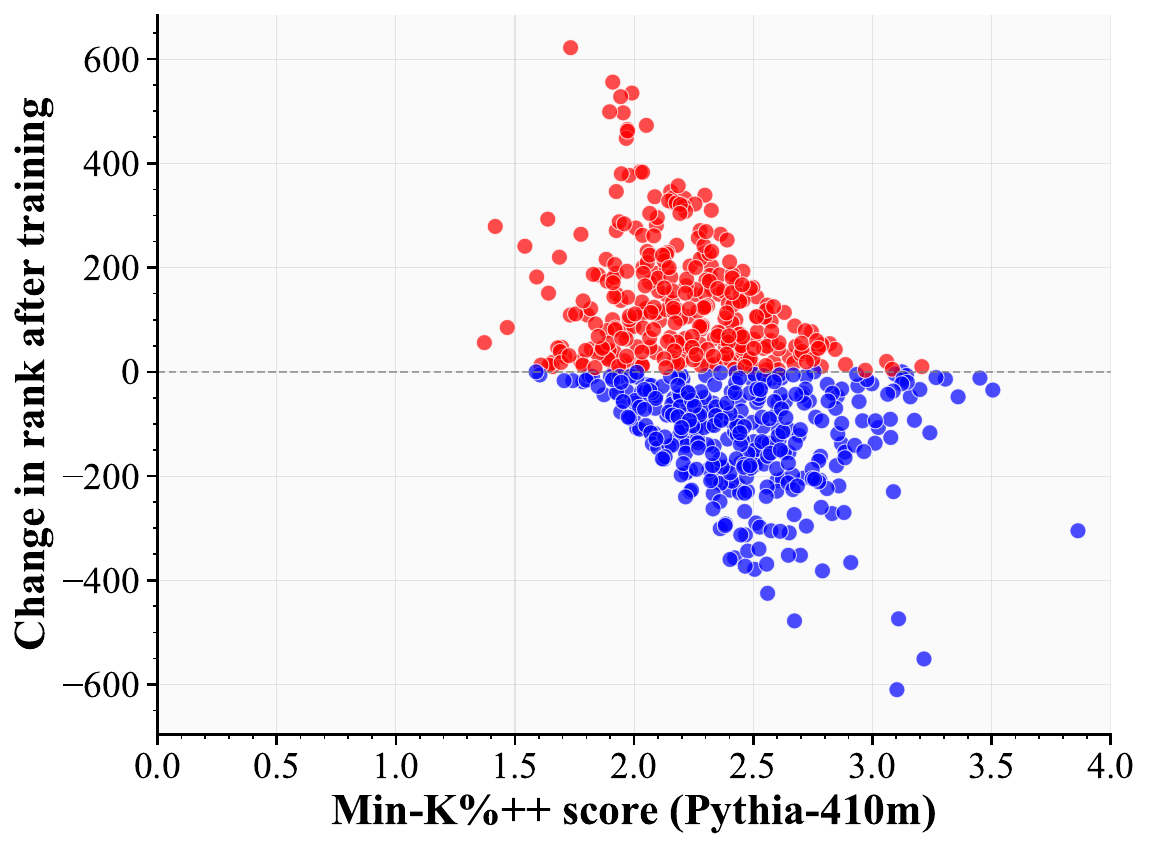}
        \caption{}
        \label{fig:subfig_b_rank_delta}
    \end{subfigure}
    
    \vspace{0.5em}
    
    \begin{subfigure}[b]{0.49\columnwidth}
        \centering
        \includegraphics[width=\textwidth]{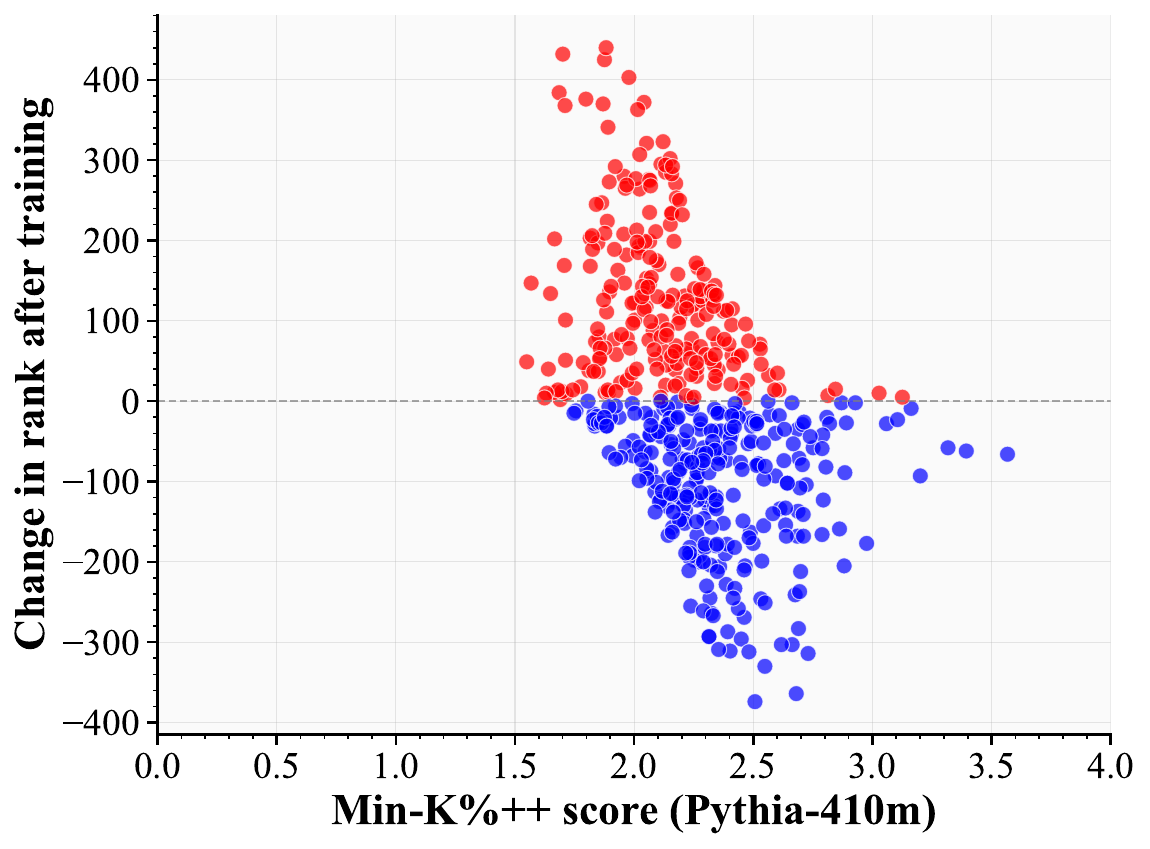}
        \caption{}
        \label{fig:subfig_c_rank_delta}
    \end{subfigure}
    \hfill
    \begin{subfigure}[b]{0.49\columnwidth}
        \centering
        \includegraphics[width=\textwidth]{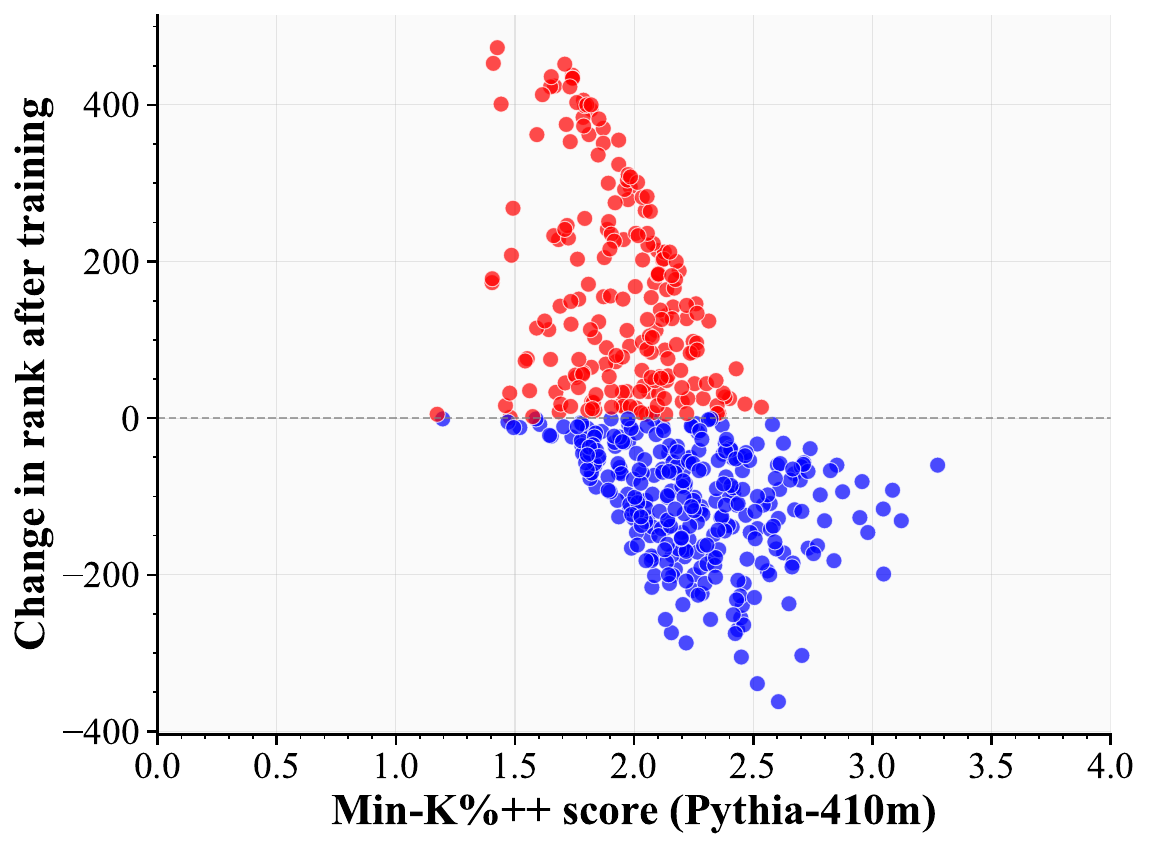}
        \caption{}
        \label{fig:subfig_d_rank_delta}
    \end{subfigure}
    \caption{Change in rank of a document when \minkpp{} is computed using Pythia-410m-CPT relative to the rank of the Pythia-410m model for different datasets a) CommonCrawl b) Reddit c) HackerNews d) PubMed}
    \label{fig:mink_rank_app}
\end{figure}

\end{document}